\documentclass[10pt,twocolumn,letterpaper]{article}

\usepackage{iccv}              % To produce the CAMERA-READY version
\usepackage[accsupp]{axessibility}  % Improves PDF readability for those with disabilities.
\DeclareMathOperator*{\argmax}{arg\,max}
\DeclareMathOperator*{\argmin}{arg\,min}
\usepackage{multirow}
\usepackage{comment}
\usepackage{algorithm}
\usepackage{algpseudocode}

\let\oldReturn\Return
\renewcommand{\Return}{\State\oldReturn}

\newcommand{\mytitle}{AdaLLaVA}
\newcommand{\tmytitle}{\texttt{\mytitle}}
\newcommand{\mytitlebold}{\textbf{AdaLLaVA}}

\usepackage{colortbl} % For coloring rows and columns
\usepackage{xcolor}   % For defining colors
\definecolor{myPurple}{RGB}{181, 27, 117}
\definecolor{myYellow}{RGB}{248, 208, 130}
\usepackage{url}

\usepackage{float}  % Add this to preamble

\definecolor{iccvblue}{rgb}{0.21,0.49,0.74}
\usepackage[pagebackref,breaklinks,colorlinks,allcolors=iccvblue]{hyperref}

\def\paperID{6637} % *** Enter the Paper ID here
\def\confName{ICCV}
\def\confYear{2025}

\title{Learning to Inference Adaptively for Multimodal Large Language Models}

\author{Zhuoyan Xu\textsuperscript{1}\thanks{Equal contribution}, Khoi Duc Nguyen\textsuperscript{1}\footnotemark[1], Preeti Mukherjee\textsuperscript{2}, \\Saurabh Bagchi\textsuperscript{2}, Somali Chaterji\textsuperscript{2}, Yingyu Liang\textsuperscript{1,3}, Yin Li\textsuperscript{1}\\
{\textsuperscript{1}University of Wisconsin-Madison \ \textsuperscript{2}Purdue University \ 
\textsuperscript{3}The University of Hong Kong}\\
}

\begin{document}
\maketitle
\begin{abstract}
Multimodal Large Language Models (MLLMs) have shown impressive capabilities in visual reasoning, yet come with substantial computational cost, limiting their deployment in resource-constrained settings. Despite recent effort on improving the efficiency of MLLMs, prior solutions fall short in responding to varying runtime conditions, in particular changing resource availability (e.g., contention due to the execution of other programs on the device). To bridge this gap, we introduce \mytitlebold, an adaptive inference framework that learns to dynamically reconfigure operations in an MLLM during inference, accounting for the input data and a latency budget. We conduct extensive experiments across benchmarks involving question-answering, reasoning, and hallucination. Our results show that \mytitle~effectively adheres to input latency budget, achieving varying accuracy and latency tradeoffs at runtime. Further, we demonstrate that \mytitle~adapts to both input latency and content, can be integrated with token selection for enhanced efficiency, and generalizes across MLLMs. Our project webpage with code release is at \href{https://zhuoyan-xu.github.io/ada-llava/}{https://zhuoyan-xu.github.io/ada-llava/}.%Moreover, our approach is compatible with other efficiency methods, such as token pruning.
\end{abstract}

% When compared to latest method, AdaInf attains a major improvement in accuracy under a wide range of latency budgets.   
\section{Introduction}
% \ZX{
% \begin{itemize}
%     \item large model inference costs.
%     \item motivation: large model pretrained with droppath. cite drop path pretrained paper.
%     \item we can even finetune the model, take any pretrained ckpt and finetune.
% \end{itemize}}

Large language models (LLMs)~\cite{openai2023gpt4,claude3} have recently been extended to connect visual and textual data, giving rise to multimodal large language models (MLLMs).
Exemplified by LLaVA~\cite{liu2023llava,liu2023improvedllava} and other works~\cite{LLaVA-NeXT,li2024llava,li2023blip,zhu2024minigpt,alayrac2022flamingo,Qwen2VL}, MLLMs have shown impressive visual reasoning capabilities, but come with significant computational costs. 
Several efforts have sought to improve the efficiency of MLLMs by exploring lightweight architectures, mixture of experts, or token selection techniques~\cite{zhu2024minigpt,yao2024minicpm,lin2024moe,shang2024llava, chen2024image}. However, prior approaches typically exhibit a fixed accuracy and latency footprint during inference, rendering them incapable of adapting to varying compute budget or input content. 
%A common characteristic of these MLLM methods is that they generate models with a fixed accuracy and latency footprint when performing inference on a given input.
% SB (8/1/25): The above statement is unclear. Do you mean that these models cannot adjust their accuracy and latency according to the complexity of the input? 
% Zhuoyan 8/1/25: Yes, these models are static, do not adjusts this based on input
% Yin: I've revised the text here to make it more clear.

% Yin: The following sentence slows down the flow, which we do not want
%Notably, token selection has become the dominant approach in the literature due to its efficiency in reducing a substantial portion of visual tokens.
% A common characteristic of these methods is that they produce models with static accuracy and latency footprint during inference.
% Yin: At this point, this is not a limitation yet
% However, a common limitation of these methods is that they yield models with fixed accuracy and latency footprints during inference.

\begin{figure}[t]
\vspace{-0.2em}
\begin{center}
{\includegraphics[width=0.75\linewidth]{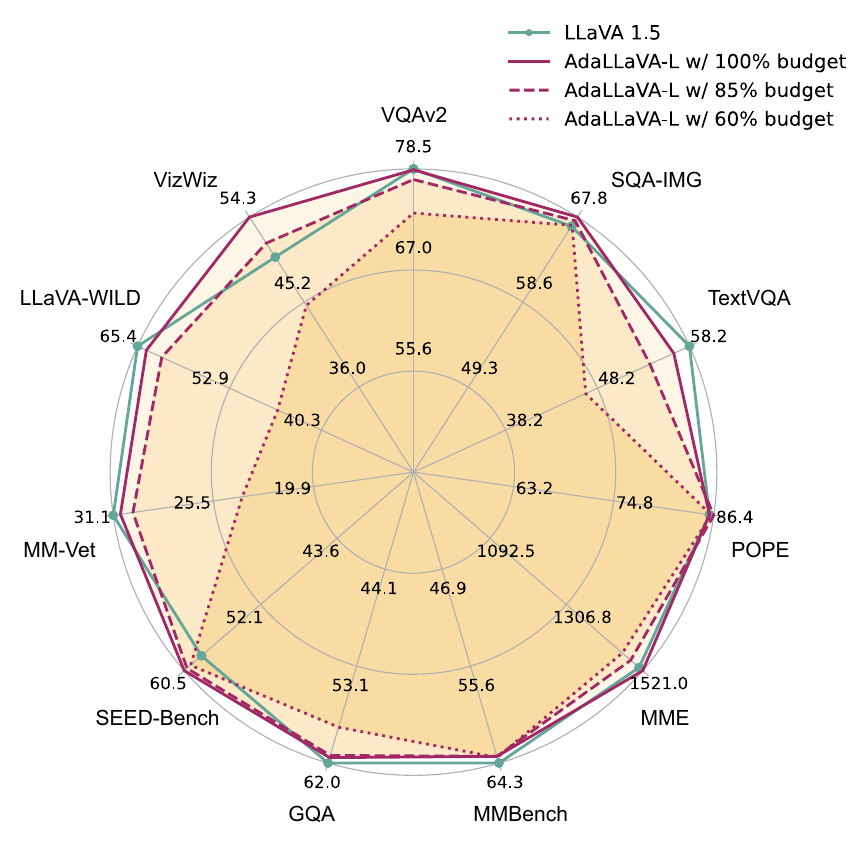}}\vspace{0.05in}
{\includegraphics[width=0.85\linewidth]{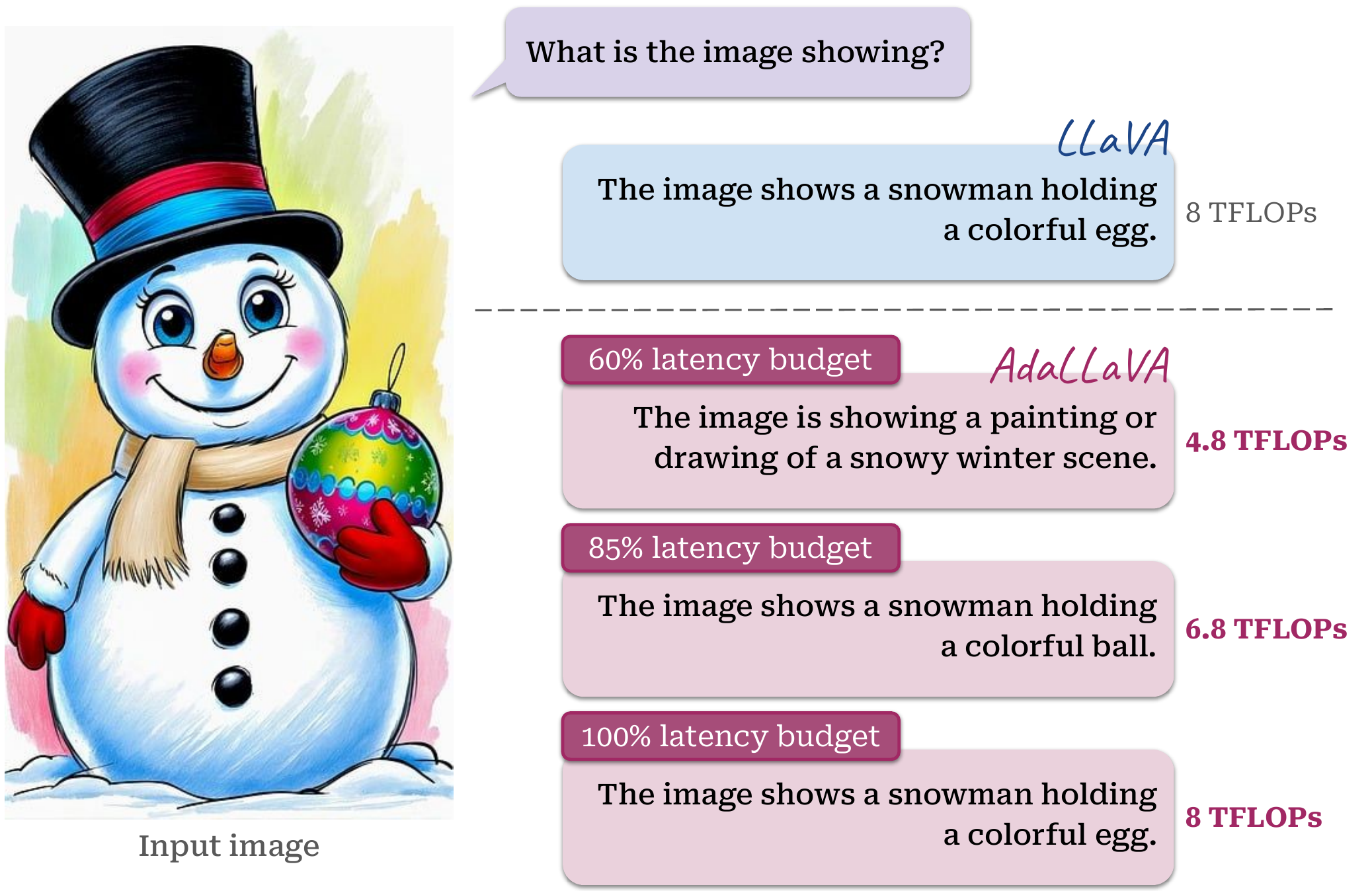}}
\end{center}
\vspace{-0.2in}
\caption{\textbf{Top}: \mytitlebold~empowers a base LLaVA model with the ability to adapt to varying compute budgets at inference time while maintaining minimal performance loss. \textbf{Bottom}: Given an image, a text query and a latency budget, \mytitle~learns to reconfigure operations within a base MLLM, generating appropriate responses while sticking to the budget.}
\vspace{-1.6em}
\label{fig:teaser}
\end{figure}

We argue that MLLMs with fixed computational footprints are insufficient for real-world deployment. 
Consider an example of deploying an MLLM on a server farm.
% Different requests may have different latency requirements, \eg, requests from a mobile application require instant feedback to users, while those from a recommendation system can tolerate higher latency due to less frequent updates. 
% SB (8/1/25): I do not understand the last part about recommendation system. If an end-user is interacting with a recommendation system (common case), then the user also wants instant feedback. Rather, you can motivate this by saying: If it is being used to create a document summarizing the content of many books, this is not an interactive task and can tolerate higher latency. 
Different requests may have different latency requirements, \eg, requests from a mobile application require instant feedback to users, while asynchronous processing tasks such as video summarization can tolerate higher latency due to their non-interactive nature.
% Zhuoyan 8/1/25: rephrased.
% Yin: minorly rephrased
Further, the available computing resources may vary over time as the overall load on the system fluctuates. 
Similarly, when deployed on an edge device, the latency budget often remains constant, yet the computing resources may vary due to contention produced by other concurrent programs.
% Yin: Added the following sentence to highlight the research challenge
In spite of this need, developing adaptive inference strategies for MLLMs that are robust across varying computational budgets~\cite{karayev2014anytime} remains an open research challenge. 

% Yin: Please update the figure caption

% Yin: revisit the legend in the top figure

To bridge this gap, we propose \textit{latency-aware adaptive inference for MLLMs}, aiming to dynamically adjust a model's computational load based on input content and a specified latency or compute budget.\footnote{In this paper, we measure a model's latency and its budget using the number of floating-point operations (FLOPs). Thus, the terms ``compute budget'' and ``latency budget'' are used interchangeably throughout.} This problem is of both conceptual interest and practical significance. Our key insight is that a modern MLLM can be viewed as a collection of shallower models, where choosing among these models enables dynamic reconfiguration during inference. For example, prior works have shown that Transformer blocks in an LLM and some attention heads within these blocks can be bypassed with minimal impact on accuracy, while reducing latency~\cite{song2024powerinfer,cai2024flextron,khaki2024the}. Therefore, strategically selecting these operations during inference results in a set of models with shared parameters but distinct accuracy-latency tradeoffs, allowing the MLLM to flexibly respond to varying latency budgets and content complexity.
% SB (8/1/25): Also varying content complexity. 
% zhuoyan 8/1/25: done.
% Yin: This sentence is out of context.
%Moreover, operation selection represents an orthogonal research direction to token selection, offering the potential to further enhance overall system efficiency.

%We have verified that these findings also extend to MLLMs, aligning with results from recent research~\cite{}. Skipping operations reduces latency.  For example, if the inference latency budget is tighter than what is need to run the full model, it is then possible to select and execute a subset of operations within the MLLM, so as to meet the latency requirement while maintaining the model's performance.

To this end, we present \mytitlebold, a learning-based framework for adaptive inference in MLLMs. As shown in \cref{fig:teaser}, given an input image, a text query, and a latency budget, \mytitle~empowers an MLLM to answer the query about the image while adhering to the specified budget --- a capability unattainable with the base MLLM. The key to \mytitle~lies in a learned scheduler that dynamically generates an execution plan, selecting a subset of operations within the MLLM based on the input content and a specified latency budget. This execution plan ensures that inference is performed within the given budget while maximizing expected accuracy.
To enable effective learning of the scheduler, we introduce a probabilistic formulation in tandem with a dedicated sampling strategy, designed to account for latency constraints at training time.

We conduct extensive experiments to evaluate \mytitle. Our results demonstrate that \mytitle~can achieve a range of accuracy-latency tradeoffs at runtime. \mytitle~exhibits strong adaptability to different latency budgets, effectively trading accuracy for compute during inference. Across several benchmarks, \mytitle~retains comparable performance to its base MLLM while operating with higher efficiency (see \cref{fig:teaser}). For example, on several comprehensive benchmarks, \mytitle~can achieve 99.0\% and 98.2\% average performance of the baseline LLaVA model when using only 80\% and 65\% of the latency budget, respectively. 
% SB (8/1/25):  We should mention here that the baseline model is LLaVA.
% Zhuoyan 8/1/25: Done.

%\yin{maybe replace this claim with the average across all datasets}
Importantly, it consistently adheres to specified latency constraints and generates content-aware execution plans tailored to input images.
Furthermore, we show that \mytitle~can be integrated with existing token selection techniques designed to enhance efficiency, making it a versatile solution for adaptive inference in MLLMs.

%\mytitle~can be integrated with token selection techniques to enhance efficiency, 

%our method demonstrates content-aware optimization by generating custom execution solutions based on specific input samples, achieving superior performance under given latency constraints compared to baseline approaches that rely on naive compression or model truncation, 
% Yin: summarize the main results / findings.
\smallskip
Our key \textbf{contributions} are summarized as follows. 
\begin{enumerate}
    % conceptually what we did, achieve adaptivity
    \item We present \mytitle, a novel adaptive inference framework for MLLMs. Our method is among the first to enable dynamic execution of MLLMs based on a latency budget and the input content at inference time.
    % technical innovation
    \item Our key technical innovation lies in (1) the design of a learning-based, latency-aware scheduler, which reconfigures a base MLLM model during inference; and (2) a probabilistic modeling approach, which incorporates hard latency constraints during MLLM training.
    % compatible to other technique
    \item Through extensive experiments, we demonstrate that (1) \mytitle~can adapt to a range of latency requirements while preserving the performance of the base model; and (2) \mytitle~can be integrated with token selection techniques to further enhance efficiency.
\end{enumerate}
\section{Related Work}

\noindent \textbf{Multimodal large language models (MLLMs)}. There has been a growing interest in extending text LLMs to multimodal signals, including images~\cite{liu2023llava}, video~\cite{li2024llava}, and audio~\cite{latif2023sparks}. This leads to the emergence of MLLMs, often involving combining vision encoders with existing LLMs. Flamingo~\cite{alayrac2022flamingo} inserts gated cross-attention dense blocks between vision encoder and LLMs to align vision and language modality. BLIP2~\cite{li2023blip} introduces Q-former with two-stage pretraining, bridging frozen image encoders and LLMs to enable visual instruction capability. LLaVA~\cite{liu2023llava,liu2023improvedllava} and MiniGPT-4~\cite{zhu2024minigpt} use a simple MLP to connect vision embedding and text token, achieving impressive performance across various tasks. Our work builds on these developments and aims to enable adaptive inference in MLLMs under varying latency budgets. 

\smallskip
\noindent \textbf{Adaptive inference}.
Adaptive inference refers to the capability in which the computational complexity of making predictions is dynamically adjusted based on the input data, latency budget, or desired accuracy level~\cite{han2021dynamic}. Early works focused on the selection of hand-crafted features in multi-stage prediction pipelines~\cite{karayev2014anytime,xu2012greedy,grubb2012speedboost}. More recent works have extended these ideas to deep models. For convolutional networks, methods have been developed to selectively downsample inputs, skip layers, or exit early during inference~\cite{figurnov2017spatially, li20212d, wang2018skipnet, bengio2015conditional, wu2018blockdrop, hu2019learning, jie2019anytime, meng2020ar,xu2022smartadapt,xu2022litereconfig}.  For vision Transformers, various approaches have been proposed to select different image patches~\cite{wang2021not, rao2021dynamicvit, pan2021ia}, or choose different attention heads and blocks~\cite{meng2022adavit,khaki2024the}. Similar ideas have also been explored for LLMs and recently MLLMs, where models selectively process tokens~\cite{raposo2024mixture,zhong2024aim} or execute a subset of the operations~\cite{du2022glam,rotem2023finding} during inference.

Our approach is conceptually similar to existing methods by dynamically selecting a subset of model components during inference. Yet unlike prior methods, our work specifically targets the latency-aware inference of MLLMs, predicting feasible execution plans tailored for input while adhering to varying latency budgets.

\smallskip
\noindent \textbf{Efficient inference for MLLMs}.
MLLMs face a major challenge in deployment due to their high computational costs during inference. Several works have designed lightweight model architectures to reduce the costs. Examples include Phi-2~\cite{javaheripi2023phi}, TinyGPT-V \cite{yuan2023tinygpt} and LLaVA-$\phi$~\cite{zhu2024llava}. 
%Microsoft released as a small LLM. Later Tinygpt-v \cite{yuan2023tinygpt} and Llava-$\phi$ \cite{zhu2024llava} integrate Phi-2 and develop small MLLM yields good performance. 
Vary-toy~\cite{wei2024small} enhances performance through specialized vision vocabulary in smaller models. TinyLLaVA~\cite{zhou2024tinyllava} and LLaVA-OneVision~\cite{li2024llava} learn small-scale models with curated training data and pipeline. MoE-LLaVA~\cite{lin2024moe} and LLaVA-MoD~\cite{shu2024llava} improve efficiency by incorporating mixture-of-experts architectures and parameter sparsity techniques. Recent works also investigate input token selection, as an input image or video can produce a large number of vision tokens. MADTP~\cite{cao2024madtp} and LLaVA-PruMerge~\cite{shang2024llava} introduce token pruning and merging technique to reduce the tokens counts. Recently, Pham et al.\ \cite{pham2024quadratic} propose to selectively disabling attention mechanisms for visual tokens in MLLMs.

While our approach also aims to improve the efficiency of MLLMs, it focuses on dynamically adjusting an MLLM to fit varying latency budgets during inference. This makes our approach orthogonal to prior efforts on developing inherently more efficient MLLMs. Through our experiments, we will demonstrate that our approach is compatible with lightweight models and integrates seamlessly with existing token-pruning techniques (\eg,~\cite{shang2024llava,chen2024image}). 

\section{Adaptive Inference of MLLMs}

\begin{figure*}
  \centering
   {\includegraphics[width=0.9\linewidth]{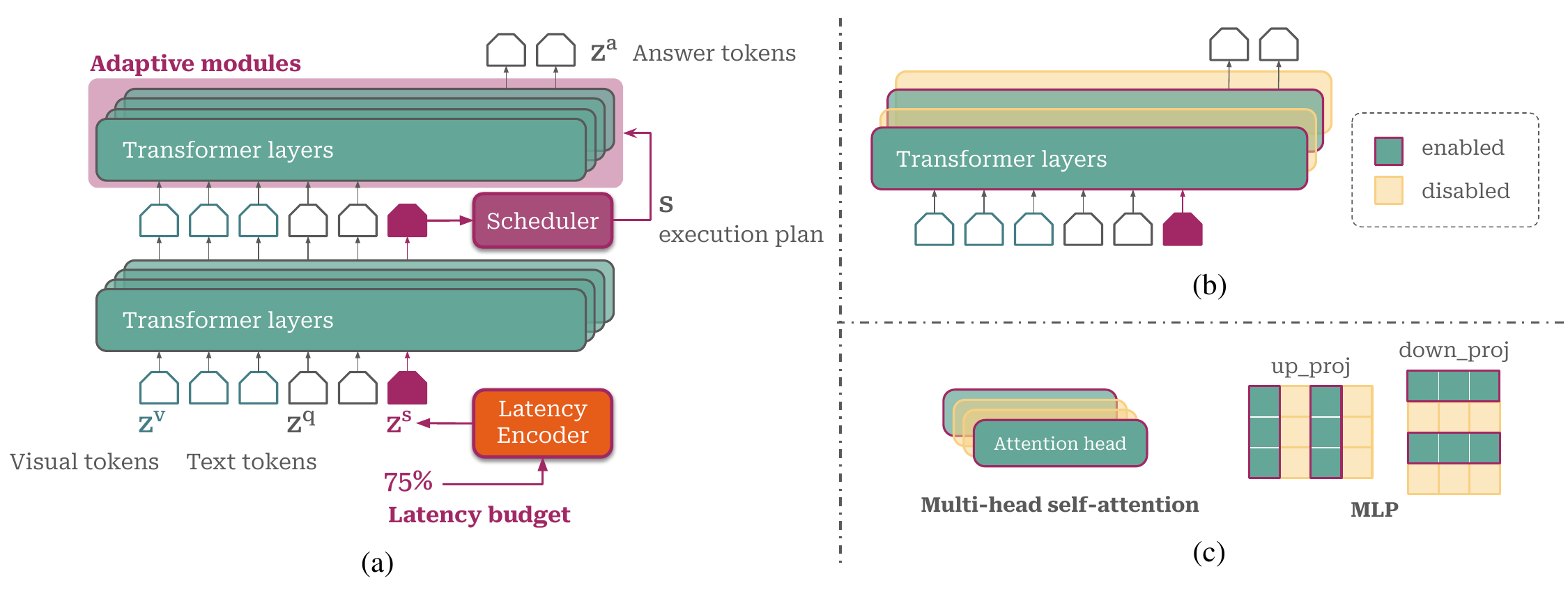}}\vspace{-1.2em}
\caption{\small Overview of \mytitlebold. \textbf{(a) Model architecture}: Our latency encoder embeds an input latency budget into a latency token, which is further processed by the early part of the LLM. The resulting embedding is then fed into the scheduler, leading to the output of an execution plan that controls individual operations in the remaining part of the LLM. Our latency encoder and scheduler are jointly trained with the MLLM.
% . This plan determines the execution of either individual Transformer blocks (\tmytitle\texttt{\small-L}) or components within the blocks (\tmytitle\texttt{\small-H}).
\textbf{(b) \tmytitle\texttt{\small-L}}: the scheduler controls the execution of entire Transformer blocks. \textbf{(c) \tmytitle\texttt{\small-H}}: the scheduler controls the execution of attention heads and MLP neurons, by masking out their activation values and the corresponding weights.}
  \vspace{-1.2em}
  \label{fig:overview}
\end{figure*}

% overivew
We now present \mytitlebold, our adaptive inference framework for MLLMs. Given a latency budget and an input image-query pair at inference time, \mytitle~leverages a scheduler learned from data to dynamically reconfigure the execution of MLLMs. Importantly, this scheduler strategically selects a subset of operations to execute, catered to the input budget and content. In doing so, \mytitle~ensures that the inference adheres to the latency constraint while preserving model accuracy. \cref{fig:overview} (a) provides an overview of our framework, where our designed scheduler takes an input of both multimodal sample and latency budget, and outputs an execution plan. In what follows, we introduce the background on MLLMs (\cref{subsec:prelim_MLLM}), outline our key idea for scheduling MLLMs (\cref{subsec:reconfig_schedule}), present our approach for training and inference with the scheduler (\cref{subec:train_infer}), and further describe the details of our solution (\cref{subsec:ModelInstantiation}).

\subsection{Preliminaries: MLLMs} \label{subsec:prelim_MLLM}
An MLLM takes an image (or video) $\mathbf{X}^v$ and a text query $\mathbf{X}^q=\{x^q\}$ as its input, and generates an answer $\mathbf{X}^a=\{x^a\}$ in text format. 
Specifically, $\mathbf{X}^v$ is first encoded by a visual encoder $h_v(\cdot)$ (including a vision backbone and its projector) into a set of visual tokens $\{\mathbf{z}^v \in \mathbb{R}^d\}$. 
Similarly, $\mathbf{X}^q$ is processed by a text encoder $h_t(\cdot)$, which embeds the words $x^q$ into a set of text tokens $\{\mathbf{z}^q \in \mathbb{R}^d\}$ with $\mathbf{z}^q = h_t(x^q) $.
These tokens are combined into $\{\mathbf{z}^{v|q}\} = [\{\mathbf{z}^v\}, \{\mathbf{z}^q\}]$, and processed by an LLM $f(\cdot)$, which decodes the answer $\mathbf{X}^a$ in an autoregressive manner:
\begin{equation}
\small
    f\left( \left[ \{\mathbf{z}^{v|q}\}, \{\mathbf{z}^a_{<i}\} \right]; \mathbf{\theta} \right) \rightarrow x_i^a, \label{eq:llm}
    % p_\theta(x_i^a|\{\mathbf{z}^{v|q}\}, \{\mathbf{z}^a_{<i}\}) = f\left( \left[ \{\mathbf{z}^{v|q}\}, \{\mathbf{z}^a_{<i}\} \right]; \mathbf{\theta} \right),
% \label{eq:llm}
\end{equation}
where $\{\mathbf{z}^a_{<i}\}$ are text tokens from previously generated answer $x^a_{<i}$, \ie $\mathbf{z}^a = h_t(x^a)$, and $\mathbf{\theta}$ denotes LLM parameters. 

For the rest of our paper, we will primarily consider the learning of LLM parameters $\mathbf{\theta}$---the major portion of parameters within the MLLM. Yet we note that learning encoder parameters (in $h_v(\cdot)$ and $h_t(\cdot)$) can be done similarly. 

\subsection{Reconfiguring and Scheduling MLLMs} \label{subsec:reconfig_schedule}
% Better move to intro. Will revisit later.

%Our key insights are that (1) MLLM can be conceptualized as a collection of shallower models, and (2) this property can be exploited to dynamically reconfigure MLLMs at inference time. 

%In particular, prior works have shown that a portion of the Transformer blocks in an LLM~\cite{}, as well as some attention heads in these blocks~\cite{}, can be bypassed with only minor drop in accuracy. We have verified that these findings also extend to MLLMs, aligning with results from recent research~\cite{}. Skipping operations reduces latency. Strategically selecting operations with varying impact to accuracy thus yields a set of models with shared parameters but distinct accuracy-latency tradeoffs, enabling dynamically reconfiguration at runtime. For example, if the inference latency budget is tighter than what is need to run the full model, it becomes possible to select and execute a subset of operations within the MLLM, so as to meet the latency requirement while maintaining the model's performance.  

\noindent \textbf{Dynamic reconfiguration}. Our key insight is that an MLLM can be conceptualized as a collection of shallower models with shared parameters, each offering a distinct accuracy-latency tradeoff.
This perspective enables dynamic reconfiguration of the MLLM during inference to meet varying latency budgets.
To this end, we propose equipping the LLM $f(\cdot)$ with $K$ tunable binary switches $\mathbf{s} \in (0, 1)^K$, which control the execution of individual operations, such as Transformer blocks or attention heads, at runtime.
Each switch determines whether a specific operation will be executed (1) or skipped (0).
% The state of each operation within a selected set is controlled by a switch, enabling (1) or disabling (0).
We defer the choice of these operations and the design of these switches to our model instantiation. Here, we focus on the concept of reconfigurable LLM decoding, expressed as
\begin{equation}
\small
\begin{split}
    f\left( \left[ \{\mathbf{z}^{v|q}\}, \{\mathbf{z}^a_{<i}\} \right], \mathbf{s}; \theta \right) \rightarrow x_i^a. 
    % p_\theta(x_i^a|\{\mathbf{z}^{v|q}\}, \{\mathbf{z}^a_{<i}\}, \mathbf{s}) = f\left( \left[ \{\mathbf{z}^{v|q}\}, \{\mathbf{z}^a_{<i}\} \right], \mathbf{s}; \theta \right). 
\end{split}
\end{equation}

Specifically, $f(\cdot)$ now takes the switches $\mathbf{s}$ as an additional input, and selectively executes a subset of operations when generating its output. 
Note that the switches $\mathbf{s}$ do not depend on the decoding step $i$, \ie, given the input tokens, a fixed set of operations is applied to generate all output tokens, although the operations may vary for different inputs.
% a fixed set of operations are applied to generate all output tokens, though this set may varying across different inputs.
% It is worth noting that our design of $\textbf{s}$ does not condition on the decoding step $i$, \ie, a fixed set of operations will be used to decode all tokens in the output, though this set may varying across different inputs. 

\smallskip
\noindent \textbf{Scheduler}. The core of our method is a scheduler $g(\cdot)$ that controls the execution of $f(\cdot)$ during inference. The scheduler $g(\cdot)$ is trained to predict a configuration of switches $\textbf{s}$ based on the input tokens $\{\mathbf{z}^{v|q}\}$ and an inference latency budget $l$. This is written as
\begin{equation}
\small
    g\left( \{\mathbf{z}^{v|q}\}, l; \mathbf{\phi} \right) \rightarrow \mathbf{s},
\end{equation}
where $\mathbf{\phi}$ denotes the parameters of the scheduler $g(\cdot)$.

The goal of $g(\cdot)$ is to determine an execution plan that meets the latency requirement while maximizing the accuracy. This requires solving the following combinatorial optimization problem \textit{for each input sample}:
\begin{equation}
\small
\begin{split}
    & \min_{\mathbf{s}} \ \ -\Sigma_i \log p\left( x^a_i=f \left( \left[ \{\mathbf{z}^{v|q}\}, \{\mathbf{z}^a_{<i}\} \right], \mathbf{s}; \theta \right) \right) \\
    & \text{s.t.} \quad \text{Latency}\left( f \left( \left[ \{\mathbf{z}^{v|q}\}, \{\mathbf{z}^a_{<i}\} \right], \mathbf{s}; \theta \right) \right) \le l.
\end{split}\label{eq:opt}
% \begin{split}
%     & \min_{\mathbf{s}} \ \ -\Sigma_i \log p_\theta(x_i^a|\{\mathbf{z}^{v|q}\}, \{\mathbf{z}^a_{<i}\}, \mathbf{s}) \\
%     & \text{s.t.} \quad \text{Latency}\left( f \left( \left[ \{\mathbf{z}^{v|q}\}, \{\mathbf{z}^a_{<i}\} \right], \mathbf{s}; \theta \right) \right) \le l.
% \end{split}\label{eq:opt}
\end{equation}
The objective here is to minimize the negative log likelihood of the target token---the standard loss used for training MLLMs, while the constraint ensures that the latency of executing the model falls within the budget. 

%Conceptually, $g(\cdot)$ examines the input of image and text query, accounts for the latency budget $l$, and decides on an execution plan $\mathbf{s}$ of the MLLM, aiming to meet the latency requirement while maintaining the accuracy. 

\subsection{Learning to Schedule Execution Plans} \label{subec:train_infer}

Learning the scheduler $g(\cdot)$ poses a major challenge. While it is tempting to pursue a fully supervised approach, in which $g(\cdot)$ is trained to predict the exact solution to~\cref{eq:opt}, doing so requires solving the optimization for each sample at every iteration during training. Even with a small number of switches, this is prohibitively expensive.

\smallskip 
\noindent \textbf{Deterministic modeling}. % Yin: describe our prior method and point out it does not enforce strict latency requirement and may lead to degraded execution plans. 
One possible solution is to solve a relaxed version of the constrained optimization at training time. We initially explored this solution, where we task $g(\cdot)$ to predict a hard execution plan with binary switches $\mathbf{s}$ and attribute latency violation as part of the objective. This leads to the following loss  
\begin{equation*}
\small 
 % \argmin_{\theta, \phi} \ -\Sigma_i \log p_\theta\left( x^a_i| \cdot \right) + \lambda \max(0, \text{Latency}(f(\cdot)) - l).
 \argmin_{\theta, \phi} \ -\Sigma_i \log p\left( x^a_i = f(\cdot)\right) + \lambda \max(0, \text{Latency}(f(\cdot)) - l),
    %\argmin_{\theta, \phi} \ -\Sigma_i \log p\left( x^a_i = f(\cdot)\right) + \lambda \| \text{Latency}(f(\cdot)) - l\|_2^2,
\end{equation*}
where $\lambda$ can be treated as the Lagrange multiplier. The execution of the LLM $f(\cdot)$ relies on the output from the scheduler $g(\cdot)$, allowing the joint optimization of $f(\cdot)$ and $g(\cdot)$. 

%This deterministic approach is further discussed in the supplement. 
We empirically found that this method fails to enforce a strict latency constraint on the scheduler and often produces suboptimal execution plans that under-utilize the available resources. We demonstrate this limitation through experimental results in \cref{subsec:efficiency_results}. % and present a further discussion in our supplement.

%To address the constrained optimization problem presented in \cref{eq:opt}, we initially explored a deterministic approach where we let the scheduler predict the hard execution plans on binary switches $\mathbf{s}$ and attribute latency violation into part of the objective. We introduce a latency loss term, measured by Mean Squared Error (MSE), to capture the deviation between actual and budgeted latency consumption. This latency loss was then integrated with the original objective function from \cref{eq:opt} using Lagrangian multiplication, resulting in a \textit{Combined-Loss} that we used to jointly optimize the LLM $f(\cdot)$ and the scheduler $g(\cdot)$. However, this method fails to enforce a strict latency constraint on the scheduler and often produces suboptimal execution plans that exceed latency limits or under-utilize the available resources. We demonstrate this limitation through experimental results in \cref{subsec:ablation}.

\smallskip
\noindent \textbf{Probabilistic modeling}. In contrast, we propose a probabilistic model to further relax the constraints, avoiding directly solving \cref{eq:opt} while stabilizing the joint training of the LLM and the scheduler. 
Specifically, we task $g(\cdot)$ to model a distribution over the choice of the switches $\mathbf{s}$, in lieu of making a hard decision:
\begin{equation}
\small
    g\left( \{\mathbf{z}^{v|q}\}, l; \mathbf{\phi} \right) \rightarrow p\left( \mathbf{s}|\{\mathbf{z}^{v|q}\}, l, \mathbf{\phi} \right). 
\end{equation}
% Yin: you will always say with "slight/minor" abuse of notation.
With slight abuse of notation, we denote $p(\mathbf{s}|\{\mathbf{z}^{v|q}\}, l, \mathbf{\phi})$ as the probability over the states $\mathbf{s}$ of $K$ binary switches given the input $\{\mathbf{z}^{v|q}\}$, latency budget $l$, and the scheduler parameters $\mathbf{\phi}$. 
Ideally, $p(\mathbf{s}|\{\mathbf{z}^{v|q}\}, l, \mathbf{\phi}) = 0$ if the execution latency of $\mathbf{s}$ exceeds the budget $l$. 

%With minor abuse of the notation, we denote $p(\mathbf{s}|\{\mathbf{z}^{v|q}\}, l, \mathbf{\phi})$ as the probability of choosing a configuration of binary switches $\mathbf{s}$ given the input $\{\mathbf{z}^{v|q}\}$, latency budget $l$, and the scheduler parameters $\mathbf{\phi}$. Ideally, $p(\mathbf{s}|\{\mathbf{z}^{v|q}\}, l, \mathbf{\phi}) = 0$ if the execution latency exceed the budget and maybe positive elsewhere. 

We now re-formulate the inference of MLLM as sampling from the following hierarchical distribution. 
\begin{equation}
\begin{split}
\small
    \mathbf{s} & \sim ~p \left( \mathbf{s}| \{\mathbf{z}^{v|q}\}, l, \mathbf{\phi} \right), \\
    x^a_i & \sim ~p \left( x^a_i|\left[ \{\mathbf{z}^{v|q}\}, \{\mathbf{z}^a_{<i}\} \right], \mathbf{s}, \theta \right).
\end{split}
\end{equation}
Conceptually, this formulation defines the following generative process: (1) the scheduler $g$ considers the input and a latency budget and outputs the conditional probability of the execution plan $p(\mathbf{s}|\{\mathbf{z}^{v|q}\}, l, \mathbf{\phi})$; (2) an execution plan $\mathbf{s}$ is then sampled from the predicted distribution without violating the latency constraint;
and (3) the plan is executed to sequentially decode $x^a_i$ and generate the answer. 

\smallskip
\noindent \textbf{Modeling $p\left( \mathbf{s}|\{\mathbf{z}^{v|q}\}, l, \mathbf{\phi} \right)$}. Our design requires that the sampled execution plan strictly adheres to the latency budget while maximizing resource utilization. To achieve this, we restrict the support of $p\left( \mathbf{s}|\{\mathbf{z}^{v|q}\}, l, \mathbf{\phi} \right)$ to the states $s$ that have exactly $k$ activated switches, where $k$ is the maximum number of switches allowed to be turned on without violating $l$. Specifically, to sample $\mathbf{s}$, $g\left( \{\mathbf{z}^{v|q}\}, l; \mathbf{\phi} \right)$ first outputs a categorical distribution over $K$ available switches. Then, $k$ switches are picked one by one without replacement, following the categorical distribution.

\smallskip
\noindent \textbf{Training loss}.
The probabilistic model allows us to directly train the LLM and the scheduler with the following loss 
\begin{equation*}
\small
     \argmin_{\mathbf{\theta}, \mathbf{\phi}} \ \mathbb{E}_{\mathcal{D}} \ \left[ -\log p \left( x^a_i | \left[ \{\mathbf{z}^{v|q}\}, \{\mathbf{z}^a_{<i}\} \right], l, \theta, \phi \right) \right],
\end{equation*}
where $\mathcal{D}$ is the data distribution approximated by the training set $(\mathbf{X}^v, \mathbf{X}^q, \mathbf{X}^a, l) \sim \mathcal{D}$. By marginalizing $\mathbf{s}$, we have 
\begin{equation}
\begin{split}
\small
    p ( x^a_i |& [ \{\mathbf{z}^{v|q}\}, \{\mathbf{z}^a_{<i}\} ], l, \theta, \phi ))= \\
    &\mathbb{E}_{p ( \mathbf{s}| \{\mathbf{z}^{v|q}\}, l, \mathbf{\phi} )} \left[ p ( x^a_i|[ \{\mathbf{z}^{v|q}\}, \{\mathbf{z}^a_{<i}\} ], \mathbf{s}, \theta )
    \right].
\end{split}
\end{equation}
% p ( x^a_i |& [ \{\mathbf{z}^{v|q}\}, \{\mathbf{z}^a_{<i}\} ], l, \theta, \phi ))= \\
%     &\mathbb{E}_{\mathbf{s} \sim p \left( \mathbf{s}| \cdot \right)} \left[ p ( x^a_i|[ \{\mathbf{z}^{v|q}\}, \{\mathbf{z}^a_{<i}\} ], \mathbf{s}, \theta ) p ( \mathbf{s}| \{\mathbf{z}^{v|q}\}, l, \mathbf{\phi} )
%     \right],
%
% $p ( x^a_i | [ \{\mathbf{z}^{v|q}\}, \{\mathbf{z}^a_{<i}\} ], l, \theta, \phi ))$ $= p ( x^a_i|[ \{\mathbf{z}^{v|q}\}, \{\mathbf{z}^a_{<i}\} ], \mathbf{s}, \theta ) p ( \mathbf{s}| \{\mathbf{z}^{v|q}\}, l, \mathbf{\phi} )$
Thus, the loss function is transformed into
\begin{equation*}
\small
    \argmin_{\mathbf{\theta}, \mathbf{\phi}} \
    \mathbb{E}_{\mathcal{D}, \mathbf{s} \sim p \left( \mathbf{s}| \cdot \right)} \left[ -\log p \left( x^a_i|\left[ \{\mathbf{z}^{v|q}\}, \{\mathbf{z}^a_{<i}\} \right], \mathbf{s}, \theta \right) \right]
    ,
\end{equation*}
where $p \left( \mathbf{s}| \cdot \right) = p \left( \mathbf{s}| \{\mathbf{z}^{v|q}\}, l, \mathbf{\phi} \right)$. 

% Optimizing the loss function for an individual sample amounts to (1) sample a feasible latency budget, (2) sample an execution plan, \ie a configuration of the switches, from the scheduler output, such that it satisfies the latency budget (3) generating the output answer using the sampled execution plan, (4) evaluate the negative log likelihood of the decoded text, and (5) compute the gradients from the loss function.

\subsection{Training and Inference}
%Moving forward, we present our approximate training and adaptive inference schemes.\smallskip

\noindent \textbf{Approximate training}. We present an approximate training scheme in the context of stochastic gradient descent (SGD). Specifically, for each training sample within a mini-batch, a latency budget $l$ is first sampled uniformly from a range of possible budgets, then an execution plan $\mathbf{s}$ is sampled from $p \left( \mathbf{s}| \{\mathbf{z}^{v|q}\}, l, \mathbf{\phi} \right)$. 
% To ensure that $\mathbf{s}$ follows the budget $l$, we propose to sample a subset of $k$ switches (out of all $K$ switches) to activate, where $k$ is the maximum number of switches allowed to turn on without violating $l$. 
With the sampled $\mathbf{s}$ guaranteed to satisfy the budget $l$, the next token $x^a_i$ can be decoded and the log-likelihood $\log p \left( x^a_i|\left[ \{\mathbf{z}^{v|q}\}, \{\mathbf{z}^a_{<i}\} \right], \mathbf{s}, \theta \right)$ (\ie, the loss) can be readily computed. Optimizing this loss requires backpropagation through the sampling process $\mathbf{s} \sim p \left( \mathbf{s}| \{\mathbf{z}^{v|q}\}, l, \mathbf{\phi} \right)$, which we approximate using the Gumbel-Softmax trick~\cite{maddison2017the,jang2017categorical}. See the supplement for more details.

% Yin: revise the details here.
% Concretely, Gumbel Softmax is applied independently to individual binary switches without replacement, until $k$ switches are selected. 

\begin{comment}
More concretely, this loss can be computed by (1) sampling a training data point and a latency budget from the training data, (2) sampling an execution plan, \ie a configuration of the switches, from the scheduler output, so that it satisfies the latency budget (3) generating the answer using the sampled execution plan, and (4) evaluating the standard negative log likelihood of the decoded text. Optimizing this loss additionally requires back propagation through the sampling process $\mathbf{s} \sim p \left( \mathbf{s}| \{\mathbf{z}^{v|q}\}, l, \mathbf{\phi} \right)$, which we approximate using the Gumbel Softmax trick~\cite{maddison2017the,jang2017categorical}.
\end{comment}

\smallskip
\noindent \textbf{Adaptive inference}. During inference, the scheduler outputs the probability $p ( \mathbf{s}| \{\mathbf{z}^{v|q}\}, l, \mathbf{\phi} )$ over possible switch configurations $\mathbf{s}$, given the input $\{\mathbf{z}^{v|q}\}$ and the latency budget $l$. In theory, decoding the answer $\mathbf{X}^a$ requires marginalizing over this distribution, which is infeasible due to the large number of configurations. In practice, we approximate the inference by selecting the most probable execution plan from the scheduler. This approximation bypasses the marginalization and thus remains highly efficient. We empirically verify its effectiveness. Formally, this approximation is given by 
\begin{equation*}
\small
\begin{split}
    {x^a_i}
    &= \argmax_{x^a_i} \ \mathbb{E}_{\mathbf{s} \sim p \left( \mathbf{s}| \cdot \right)} \left[p \left( x^a_i|\left[ \{\mathbf{z}^{v|q}\}, \{\mathbf{z}^a_{<i}\} \right], \mathbf{s}, \theta \right) \right] \\
    % &= \argmax_{x^a_i} \ \mathbb{E}_{\mathbf{s} \sim p \left( \mathbf{s}| \cdot \right)} \left[ -\log p \left( x^a_i|\left[ \{\mathbf{z}^{v|q}\}, \{\mathbf{z}^a_{<i}\} \right], \mathbf{s}, \theta \right) \right] \\
    &\approx \argmax_{x^a_i} \ p \left( x^a_i|\left[ \{\mathbf{z}^{v|q}\}, \{\mathbf{z}^a_{<i}\} \right], \mathbf{s}^*, \theta \right),
\end{split}
\end{equation*}
where $\mathbf{s}^* = \argmax_{\mathbf{s}} p \left( \mathbf{s}| \{\mathbf{z}^{v|q}\}, l, \mathbf{\phi} \right)$. 
% Note that model parameters $\theta$ and $\phi$ are now fixed.

\subsection{Model Instantiation} \label{subsec:ModelInstantiation}

\noindent \textbf{Design of tunable switches}. We consider attaching binary switches to the LLM part of an MLLM, which accounts for the majority of computational costs. We explore two different designs of switches to select operations. %allowing executing or skipping these operations at inference time. 
\begin{itemize}
    \item \tmytitle\texttt{\small-L} (layer-level): This design attaches binary switches to entire Transformer blocks. When a switch is off, the corresponding block is bypassed through its residual connection, becoming an identity mapping. The execution plan thus determines whether each layer is computed or bypassed (see \cref{fig:overview}(b)).
    \item \tmytitle\texttt{\small-H} (head/neuron-level): This design introduces binary switches within Transformer blocks, targeting individual attention heads in attention modules and specific neurons in MLP layers. When a switch is off, its computation is skipped, and its contribution is removed. In MLP, switches function similarly to dropout~\cite{srivastava2014dropout}, selectively disabling neuron activations (see \cref{fig:overview}(c)). %\tmytitle\texttt{\small-H} enables finer control over the accuracy-latency tradeoff compared to its layer-level counterpart .
\end{itemize}

%fixed the first half of transformer blocks, applying dynamic execution plans exclusively to the latter half.

% consider top Transformer blocks and their attention heads, and attach binary switches to them,  Please provide sufficient details

\smallskip
\noindent \textbf{Model architecture}. Our goal is to design a lightweight scheduler that minimizes computational overhead yet remains expressive enough to support effective decision-making. To this end, we reuse part of the LLM $f(\cdot)$ to extract visual-language features and encode the latency constraint for the scheduler. Specifically, we first design a latency encoder that converts a latency budget into a token embedding, which is then appended to the original input sequence before being processed by the LLM layers.
Within the LLM, the latency token is processed by a few Transformer blocks, attending to all visual-language tokens. 
%allowed to attend to the prompt visual-language tokens, making it a representative input for the scheduler.
The processed token is then passed to a lightweight scheduler that generates the execution plan for the rest of the LLM. Notably, the first few Transformer blocks in the LLM serve two purposes: it simultaneously processes regular MLLM tasks and learns resource allocation based on both content and budget constraints. This design is depicted in~\cref{fig:overview} (a).
% Representations of this latency token are extracted from intermediate layers
% attach an additional input latency token as part of the input .... Details.  

\begin{comment}
\smallskip
\noindent \textbf{Modeling $p \left( \mathbf{s}| {\mathbf{z}^{v|q}}, l, \mathbf{\phi} \right)$}. 
To model the conditional distribution $p \left( \mathbf{s}| {\mathbf{z}^{v|q}}, l, \mathbf{\phi} \right)$, our scheduler first outputs a categorical distribution over the $K$ switches.
Given the input latency constraint, we sample a subset of $k$ switches to activate, where $k$ is the maximum number of switches that can be activated without violating the constraint.
% The resulting execution plan is then implemented, and we optimize the model by minimizing our loss. %the divergence between the LLM's generated text and the ground truth responses.
\end{comment}

% Our schedule output a probability for each switch. To approximate $p \left( \mathbf{s}| \{\mathbf{z}^{v|q}\}, l, \mathbf{\phi} \right)$ (distribution of switch configurations under latency constraint), we given the output logits for each switch output by the scheduler, we implemented conditional sampling technique, where we sample switches one by one without replacement, until it meets the latency budget. We then execute the plans within the latency budget and minimize the loss between text generated by LLM and ground truth text.

\begin{table*}[t]
\centering
\scalebox{0.85}{
\begin{tabular}{llccc|ccc|ccc}
\toprule
\multirow{2}{*}{Method} & \multirow{2}{*}{LLM} & Budget & FLOPs  & Prefill time  & VQA\textsuperscript{v2} & SQA\textsuperscript{I} & VQA\textsuperscript{T} & POPE & MME & MMBench \\
 & & (\%) & (T) & (ms) & \cite{goyal2017vqav2} & \cite{lu2022learn} & \cite{singh2019towards} & \cite{li2023pope} & \cite{fu2023mme} & \cite{liu2025mmbench}\\
\midrule
% BLIP-2~\cite{li2023blip} & Vicuna-13B  & - & - & - & 41.0 & 61 & 42.5 & 85.3 & 1293.8 & - \\
% InstructBLIP~\cite{dai2023instructblip} & Vicuna-7B & - & - & - & - & 60.5 & 50.1 & - & - & 36.0 \\
% InstructBLIP~\cite{dai2023instructblip} & Vicuna-13B & - & - & - & - & 63.1 & 50.7 & 78.9 & 1212.8 & - \\
% Shikra~\cite{chen2023shikra} & Vicuna-13B & - & - & - &  77.4 & - & - & - & - & 58.8 \\
% IDEFICS-9B~\cite{IDEFICS} & LLaMA-7B & - & - & - &  50.9 & - & 25.9 & - & - & 48.2 \\
% IDEFICS-80B~\cite{IDEFICS} & LLaMA-65B & - & - & - &  60.0 & - & 30.9 & - & - & 54.5 \\
% Qwen-VL~\cite{Qwen-VL} & Qwen-7B & - & - & - & 78.8 & 67.1 & 63.8 & - & - & 38.2 \\
% Qwen-VL-Chat~\cite{Qwen-VL} & Qwen-7B & - & - & - & 78.2 & 68.2 & 61.5 & - & 1487.5 & 60.6 \\
% \hline
LLaVA-1.5~\cite{liu2023improvedllava} & Vicuna-7B & 100 & 8.6 & 81 & 78.5 & 66.8 & 58.2 & 85.9 & 1510.7 & 64.3 \\
\rowcolor{myPurple!30}
w/ \tmytitle \texttt{\small-L} & Vicuna-7B & 100 & 8.6 & 81 & 78.4 & 67.8 & 57.0 & 85.9 & 1521.0 &  63.7 \\
\rowcolor{myPurple!30}
w/ \tmytitle \texttt{\small-L} & Vicuna-7B & 85 & 7.2 & 69 & 77.1 & 67.4 & 54.5 & 86.4 & 1487.2 &  63.7 \\
\rowcolor{myPurple!30}
w/ \tmytitle \texttt{\small-L} & Vicuna-7B & 60 & 5.1 & 49 & 75.0 & 66.9 & 47.7 & 86.1 & 1463.8 & 63.8 \\
\rowcolor{myYellow!30}
w/ \tmytitle \texttt{\small-H} & Vicuna-7B & 100 & 8.6 & 81 & 77.9 & 68.5 & 57.1 & 86.9 & 1471.1 & 64.1 \\
\rowcolor{myYellow!30}
w/ \tmytitle \texttt{\small-H} & Vicuna-7B & 85 & 7.2 & 69 & 76.8 & 68.2 & 55.2 & 86.7 & 1494.9 & 64.3 \\
\rowcolor{myYellow!30}
w/ \tmytitle \texttt{\small-H} & Vicuna-7B & 60 & 5.1 & 49 & 74.2 & 68.1 & 48.7 & 85.0 & 1489.6 & 64.8 \\
% \hline
% Prumerge 7B & 100 & 1.9 & 72.0 & 68.5 & 56.0 & 76.3 & 1350.3 & 60.9 \\
% \rowcolor{myPurple!30}
% w/ \tmytitle \texttt{\small-L} & 100 & 1.9 & 71.0 & 69.1 & 54.1 & 74.2 & 1312.6 & 58.4 \\
% \rowcolor{myPurple!30}
% w/ \tmytitle \texttt{\small-L} & 85 & 1.6 & 69.7 & 68.6 & 52.5 & 75.6 & 1313.3 & 59.1 \\
% \rowcolor{myPurple!30}
% w/ \tmytitle \texttt{\small-L} & 60 & 1.1 & 67.8 & 68.7 & 44.7 & 75.8 & 1332.5 & 57.0 \\
% \rowcolor{myYellow!30}
% w/ \tmytitle \texttt{\small-H} & 100 & 0.91 & == & 67.9 & 54.4 & 77.2 & 1311.4 & 60.1 \\
% \rowcolor{myYellow!30}
% w/ \tmytitle \texttt{\small-H} & 85 & 0.77 & == & 67.2 & 52.3 & 75.5 & 1309.7 & 60.7 \\
% \rowcolor{myYellow!30}
% w/ \tmytitle \texttt{\small-H} & 60 & 0.54 & == & 68.1 & 45.9 & 76.4 & 1289.3 & 58.7 \\
\hline
Prumerge+~\cite{shang2024llava} & Vicuna-7B & 100 & 3.0 & 29 & 76.8 & 68.3 & 57.1 & 84.0 & 1462.4 & 64.9 \\ 
\rowcolor{myPurple!30}
w/ \tmytitle \texttt{\small-L} & Vicuna-7B & 100 & 3.0 & 29 & 76.3
& 68.3 & 55.8 & 85.1 & 1455.5 & 61.9 \\
\rowcolor{myPurple!30}
w/ \tmytitle \texttt{\small-L} & Vicuna-7B & 85 & 2.6 & 24 & 75.3 & 68.5 & 52.9 & 85.7 & 1429.5 & 62.5 \\
\rowcolor{myPurple!30}
w/ \tmytitle \texttt{\small-L} & Vicuna-7B & 60 & 1.8 & 17 & 73.0 & 67.7 & 47.4 & 85.6 & 1450.9 & 61.3 \\
\rowcolor{myYellow!30}
w/ \tmytitle \texttt{\small-H} & Vicuna-7B & 100 & 3.0 & 29 & 76.0 & 67.9 & 56.0 & 86.6 & 1503.2 & 63.2 \\
\rowcolor{myYellow!30}
w/ \tmytitle \texttt{\small-H}& Vicuna-7B & 85 & 2.6 & 24 & 75.0 & 68.1 & 54.2 & 86.4 & 1511.8 & 63.6 \\
\rowcolor{myYellow!30}
w/ \tmytitle \texttt{\small-H}& Vicuna-7B & 60 & 1.8 & 17 & 72.2 & 67.6 & 47.2 & 86.4 & 1458.0 & 63.6 \\

% \hline
% LLaVA-1.5 13B & 100 & 18.2 & 80.0 & 71.6 & 61.3 & 85.9 & 1531.3 & 67.7 \\
% \rowcolor{myPurple!30}
% w/ \tmytitle \texttt{\small-L} & 100 & 18.2 & 79.4 & 72.4 & 59.9 & 86.9 & 1559.3 & 69.2 \\
% \rowcolor{myPurple!30}
% w/ \tmytitle \texttt{\small-L} & 85 & 15.9 & 79.1 & 72.4 & 58.0 & 86.2 & 1563.9 & 68.9 \\
% \rowcolor{myPurple!30}
% w/ \tmytitle \texttt{\small-L} & 60 & 11.3 & 77.1 & 71.8 & 54.3 & 87.3 & 1552.6 & 68.6 \\
% \rowcolor{myYellow!30}
% w/ \tmytitle \texttt{\small-H} & 100 & 18.2 & -=+ & 72.6 & 59.9 & 87.3 & 1531.9 & 67.4 \\
% \rowcolor{myYellow!30}
% w/ \tmytitle \texttt{\small-H} & 85 & 15.9 & -=+ & 72.3 & 59.0 & 86.1 & 1554.5 & 67.0 \\
% \rowcolor{myYellow!30}
% w/ \tmytitle \texttt{\small-H} & 60 & 11.3 & +-= & 71.3 & 53.3 & 85.0 & 1529.5 & 66.9 \\
\hline
FastV (K=2,R=0.5)~\cite{chen2024image} & Vicuna-7B & 100 & 4.9 & 47 & 77.7 & 68.7 & 58.1 & 82.5 & 1516.2 & 64.3\\ 
\rowcolor{myPurple!30}
w/ \tmytitle \texttt{\small-L} & Vicuna-7B & 100 & 4.9 & 47 & 77.8 & 67.7 & 57.0 & 82.8 & 1494.3 & 63.5 \\
\rowcolor{myPurple!30}
w/ \tmytitle \texttt{\small-L} & Vicuna-7B & 85 & 4.2 & 40 & 76.9 & 67.8 & 54.4 & 83.3 & 1478.1 & 63.7\\
\rowcolor{myPurple!30}
w/ \tmytitle \texttt{\small-L} & Vicuna-7B & 60 & 3.0 & 29 & 74.5 & 67.0 & 47.4 & 83.8 & 1463.1 & 63.2\\
\rowcolor{myYellow!30}
w/ \tmytitle \texttt{\small-H} & Vicuna-7B & 100 & 4.9 & 47 & 77.4 & 68.4 & 57.0 & 84.3 & 1484.2 & 63.8 \\
\rowcolor{myYellow!30}
w/ \tmytitle \texttt{\small-H} & Vicuna-7B & 85 & 4.2 & 40 & 76.6 & 67.7 & 54.8 & 83.9 & 1520.5 & 63.9 \\
\rowcolor{myYellow!30}
w/ \tmytitle \texttt{\small-H} & Vicuna-7B & 60 & 3.0 & 29 & 73.9 & 68.3 & 48.7 & 82.4 & 1452.8 & 65.3 \\
\bottomrule
\end{tabular}
} \vspace{-0.5em}
\caption{\textbf{Results on MLLM benchmarks}. Budget (\%): input latency budget w.r.t. the base model latency. \tmytitle\texttt{\small-L}: switches on selecting different Transformer blocks. \tmytitle\texttt{\small-H}: switches on select different attention heads and MLP activations. VQA\textsuperscript{v2}: VQAv2 set. SQA\textsuperscript{I}: ScienceQA set. VQA\textsuperscript{T}: TextVQA set. Prumerge+ and FastV both use LLaVA 1.5. \mytitle~enables a base MLLM to adapt to varying latency budgets with competitive performance, and can be further integrated with token selection to enhance overall efficiency.} 
\vspace{-0.2in}
\label{tab:main_table}
\end{table*}

\smallskip
\noindent \textbf{Implementation details}. 
% All other implementation details that are not tied to a particular experiment go here. 
%In most of our experiments, we adopt the architecture of LLaVA~\cite{liu2023improvedllava} and integrate the scheduler into its LLM (see ~\cref{fig:overview}).
Our \textit{latency encoder} uses the sinusoidal positional encoding~\cite{waswani2017attention} to map the scalar latency $l$ to a 256-D vector.  A two-layer MLP, with GELU and layer norm, then converts this vector to a latency token $\mathbf{z}^s$, ready to be appended to the input sequence of the LLM (see \cref{fig:overview}(a)). Our \textit{scheduler} is implemented as a linear layer that maps the processed latency token (from the bottom part of the LLM Transformer blocks) to logits, defining a categorical distribution over switch selection. We use \textit{FLOPs} to quantify the theoretical latency budget, following the calculation in~\cite{yuan2024llm}. Specifically, we report the average prefill FLOPs on a target dataset, isolating it from variations in decoding length to ensure a more consistent evaluation.

% To ensure stable model performance, we divide all Transformer blocks into two parts, based on their order of execution in the LLM. We fix the bottom part (the first half of blocks by default), and apply tunable switches exclusively to the latter part. 
% We implement a group-wise attention head selection strategy. Specifically, attention heads are divided into groups of 8, where each group operates as a single unit. 

We split the LLM evenly into two parts unless otherwise specified. We use the first part to process the latency token, and apply tunable switches exclusively to the latter part. In AdaLLaVA-H, we attach a switch to each attention head in the self-attention. For the MLP, channels are grouped to match the number of attention heads, with each group controlled by a single switch
%\yin{I don't get the original text here.}
%For efficient execution plan sampling, the number of activated switches in each Transformer block is quantized into a multiple of 8.
This implementation reduces the design space while preserving control granularity. See ablation study on group size in the supplement. 

%Our latency encoder uses the standard sinusoidal positional encoding to map a scalar latency $l$ to $\gamma \in \mathbb{R}^{256}$. An MLP then converts $\gamma$ to a latency token $\mathbf{z}^s$, ready to be appended to the input sequence of the LLM (see \cref{fig:overview}(a)).
% Our latency encoder is implemented using sine and cosine functions, and the scheduler is a simple linear layer (randomly initialized) that maps the latency token to the logits for the switches. 
%The scheduler is a single linear layer that maps the processed latency token to a categorical distribution over the switches.

% The scheduler takes the latency token processed by the first half of the transformer layers and generates the execution plan for the second half. This results in a model with latency ranging from 50\% to 100\% of the original LLaVA model.

%\smallskip
%\noindent \textbf{Latency budget as FLOPs}. 
%To quantify model latency, we adopt a computational complexity-based approach, using FLOPs (floating point operations) as our primary metric. This provides a hardware-agnostic measure of computational cost that directly correlates with actual runtime performance. 

% \input{table/main_tab}
\section{Experiments and Results}\label{sec:exps}

We now present our experiments and results. We introduce our setup (\cref{subsec:exp_dataset_model}), present our main results (\cref{subsec:main_result}), and provide further analyses (\cref{subsec:efficiency_results}).
%, and conduct ablation studies (\cref{subsec:ablation}). 
Additional experiments, including further ablations, are included in our supplement.

%In this section, we provide our experimental setup and results. Our experimental setup includes the dataset and model we used for empirical experiments, as shown in \Cref{subsec:exp_dataset_model}. We provide our main results when apply \mytitlebold on LLaVA type MLLMs over several benchmarks in \Cref{subsec:main_result}. We then show the efficiency improvement where our methods achieves a accuracy-latency trade offs among latency requirement in \Cref{subsec:efficiency_results}. Then, we then show ablation studies in \Cref{subsec:ablation} exploring the training strategy and model design.

\subsection{Experimental Setup} \label{subsec:exp_dataset_model}
\noindent \textbf{Experiment protocol}. In most of our experiments, we build on LLaVA-1.5~\cite{liu2023improvedllava}. Training LLaVA and many other MLLMs typically involves two stages: (1) vision-language alignment pre-training; and (2) visual instruction tuning. We focus on the second stage and seek to jointly finetune the LLM within the MLLM and train our scheduler using visual instruction data, while keeping the vision encoder frozen. Once trained, we perform zero-shot inference across multiple benchmarks following the common practice in the community~\cite{liu2023improvedllava}, yet under varying latency budgets. \smallskip

\noindent \textbf{Training details}.
%Instead of following LLaVA's two-stage training procedure, we focus on jointly finetuning its LLM and training the scheduler on visual instruction data, while keeping the vision encoder frozen.
Our model is initialized with the pretrained LLaVA-1.5 checkpoints. During finetuning, each training sample is paired with a randomly sampled latency budget ranging from 0.5 to 1.0, as by default we only operate on the top half of the layers in LLM. We set the learning rate to $10^{-5}$ for the original LLaVA model and the scheduler, while keeping other training hyperparameters consistent with the original LLaVA stage-2 finetuning protocol.\smallskip
% \noindent \textbf{Evaluation datasets and metrics}. 
% During the evaluation, we tested our model under various benchmarks, with each test setting applying a consistent latency constraint (ranging from 50\% to 1) across all samples.

% Yin: explain why the latency constraint ranges from 0 to 1
% Khoi: it is explained in the implementation details.

\noindent \textbf{Benchmarks and metrics}.
We conduct a comprehensive evaluation across multiple visual understanding benchmarks, including VQAv2~\cite{goyal2017vqav2}, ScienceQA~\cite{lu2022learn}, TextVQA~\cite{singh2019towards}, MME~\citep{fu2023mme}, and MMBench~\citep{liu2025mmbench}. We also evaluate on hallucination benchmarks such as POPE~\citep{li2023pope}. For TextVQA, we specifically focus on the image-based subset, where each question is paired with its corresponding image content. For each benchmark, we report the official metrics on the same dataset splits as in LLaVA-1.5. We report accuracy for VQAv2, ScienceQA, TextVQA and MMBench, perception score for MME, and F1 score for POPE.
% \yin{Please specify the exact metrics. E.g., Specifically, we report accuracy for X, Y, and Z, and XX for ...} 
Additionally, we consider varying latency budgets (from 0.5 to 1.0) when evaluating \mytitle. We report the Prefill FLOPs and time on MME benchmark. 
%\yin{Need to briefly mention that our FLOPs are computed on MME.}
\smallskip 

\noindent \textbf{Baselines and model variants}.
% Baselines in our main results include BLIP2~\citep{li2023blip,dai2023instructblip}, Shikra~\citep{chen2023shikra}, IDEFICS~\cite{IDEFICS}, Qwen-VL~\cite{Qwen-VL}, and LLaVA-1.5~\cite{liu2023improvedllava}. Our closest competitor is the base model LLaVA-1.5~\cite{liu2023improvedllava}. 
We mainly compare our model with base model LLaVA-1.5~\cite{liu2023improvedllava}. 
% \yin{Describe the main baselines and model variants here. E.g., Our baselines include the recent works: XX, YY, ZZ... Our close competitor is the base model LLaVA-1.5.} 
We evaluate \mytitle~with 7B and 13B (see supplement) models, and across two different designs: (a) \mytitle-L for selecting Transformer blocks; and (b) \mytitle-H for selecting attention heads and MLP activations. In our additional analyses, we also use Mipha-3B~\cite{zhu2024comprehensive} as the base model. 
%\zxu{consider adding 13b-L to main table}
%To demonstrate the efficacy of \mytitle, we consider two latency budgets: 60\% and 85\%. Additionally, we report the FLOPs during the prefill stage for an efficiency comparison.

%In each evaluation of \mytitle, the same latency requirement (from 0.5 to 1.0) is applied across all sample in the dataset.

\subsection{Main Results} \label{subsec:main_result}

\begin{figure}[t!]
    \centering
    % \vspace{-0.13in}
    {\includegraphics[width=\linewidth]{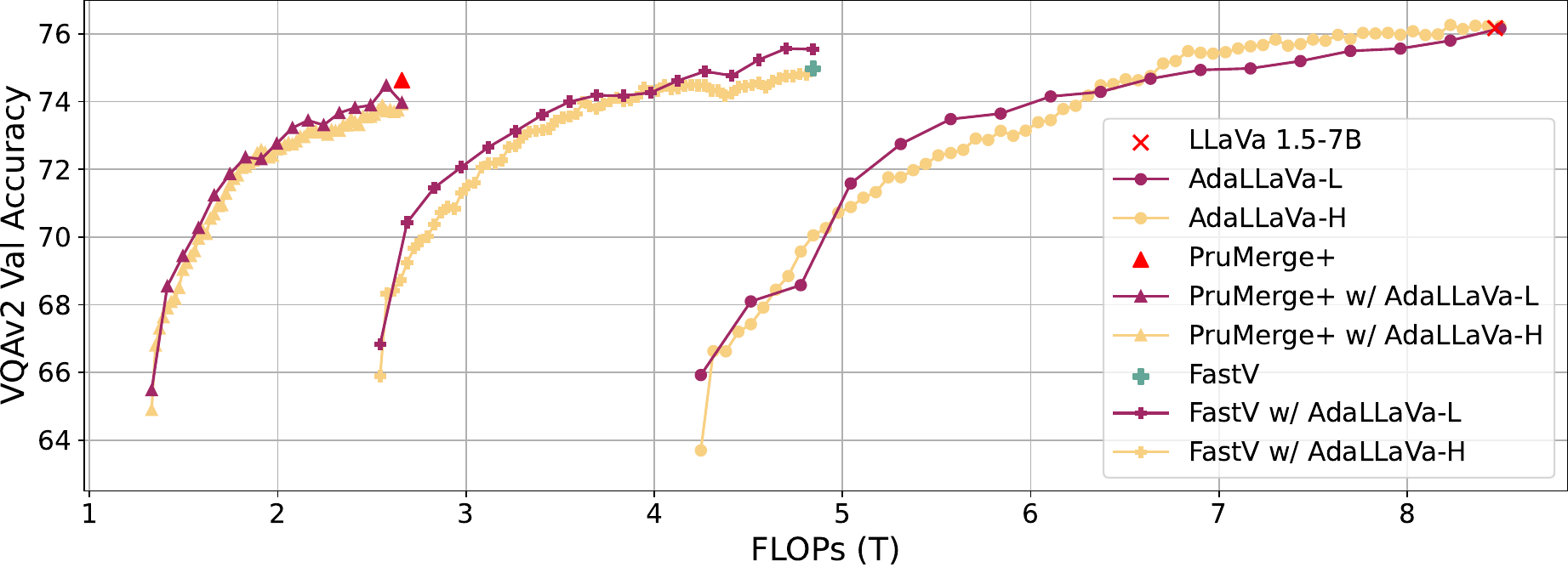}}
    \label{fig:binomial_7b_curve}
    \vspace{-2em}
    \caption{\textbf{Accuracy-latency tradoffs of \mytitle} with LLaVA-1.5-7B and additional token selection techniques (PruMerge+ / FastV). Results reported on VQAv2.}
    \vspace{-1.5em}
    \label{fig:binomial_7b_curve_with_fastv}
\end{figure}

%In this section, we present main results of \mytitlebold on several benchmarks comparing with related MLLMs, showing advantages in performance and efficiency. We show results on six widely used benchmarks we setup in \Cref{subsec:exp_dataset_model}. We apply our \mytitle~ with different design of switches (either selecting model blocks or attention heads and MLP activations) on llava-1.5 model. We conduct our methods on both 7B and 13B models, and show comparable results with our different latency input. To expand our experiments on wide range of models and benchmarks, we subset our latency input to be 60\% and 85\%, where the full model is 100\%.

%\smallskip % Yin: need to consolidate the text below
\noindent \textbf{Comparison to baselines}. Our main results across six benchmarks are summarized in \cref{tab:main_table}. \mytitle~demonstrates competitive performance with notable efficiency improvements across all benchmarks. 

\textit{\mytitle-L}, when applied to LLaVA-1.5 7B, maintains comparable performance under full computational budgets. With reduced compute budgets, \mytitle-L shows minimal performance degradation with an average accuracy drop of only 1.5\% at 85\% budget and 3.4\% at 60\% budget.  
Remarkably, at 60\% compute budget, \mytitle-L even has slightly better results than the base model on ScienceQA (66.9 vs.\ 66.8) and POPE (86.1 vs.\ 85.9). %These improvements suggest that our finetuning methodology particularly enhances the model's capabilities on these specific tasks.

\textit{\mytitle-H} shows similar results, with only 1\% average performance drop at 85\% budget, and 1.9\% at 60\% budget. The superior performance of \mytitle-H compared to \mytitle-L can be attributed to its head/neuron-level switching mechanism, allowing for more fine-grained control over computational resources than layer-level switches used in \mytitle-L. 

Importantly, for all results, \mytitle~adheres to the specified latency budgets (see  \cref{subsec:efficiency_results}). We provide results on additional VQA benchmarks in \cref{supp:sec:full_table} \cref{tab:more_task}. 
% \zxu{may remove this} Across all datasets, \mytitle~shows a larger performance drop on TextVQA, which focuses on OCR ability and reason over text in images. We conjecture that this particular task may require a model to preserve subtle visual details related to textual information, and such information may have been dropped by \mytitle. 

%\yin{On $VQA^T$, our methods have a major performance drop. Any explaination?}
%We notice \mytitle~ experiences a performance drop on TextVQA set, which focus on OCR ability and reason over texts. These tasks are more sensitive to details and need to focus on fine-grained region. Our methods limits model attention to certain image regions regions may affect the performance. 

% We present a detailed analysis of differences in our ablation studies (\cref{subsec:ablation}).

%A full set of results can be found in~\cref{supp:sec:full_table}.

% \mytitle-H achieves higher performance than \mytitle-L as head/neuron-level switches providing moregranular control over computational resources compared to the layer-level switches,

% The effectiveness of our approach also scales to larger models, as demonstrated by the results with Vicuna-13B backbone. Notably, in some cases, our method outperforms the baseline while using only 60\% computational resources, as seen in ScienceQA (71.9 vs 71.6), POPE (86.9 vs 85.9) and MMB (68.5 vs 67.7). 

\smallskip 
\noindent \textbf{Integration with token selection}. Token selection techniques have demonstrated recent success in improving the efficiency of MLLMs~\cite{shang2024llava,chen2024image}. \mytitle~presents an orthogonal direction in adaptive inference. We now demonstrate that \mytitle~can be integrated with token selection to further enhance the efficiency. We combine \mytitle~with PruMerge+~\cite{shang2024llava} and FastV~\cite{chen2024image}, two latest token selection methods designed for MLLMs. For FastV, we set filtering layer $K$=2 and filtering ratio $R$=50\% to ensure consistent comparison. The results are shown in~\cref{tab:main_table}.

%Our method demonstrates strong compatibility with other efficiency techniques, such as token pruning method. We integrate \mytitle~with integrated with LLaVA-PruMerge \cite{shang2024llava} and FastV \cite{chen2024image}. For FastV, we used fixed parameters of filtering layer $K=2$ and filtering ratio $R=5\%$ to ensure consistent comparison.

With the integration of PruMerge+ or FastV, \mytitle\ shows significantly improved efficiency across board, when compared to \mytitle\ with LLaVA-1.5. Again, \mytitle~adapts to varying latency budgets and achieves competitive performance relative to the base model (PruMerge+/FastV). For example, with PruMerge+, \mytitle-H shows 2.45\% average performance boost at 85\% compute budget and only 1.01\% performance drop at 60\%. %\yin{Fill in the number here.}
A surprising observation is that \mytitle-H achieves strong performance at 85\% latency budget, sometimes beating the base model with token pruning.
%This suggests our approach-H performs particularly well when integrated with token pruning methods. 
Overall, our results suggest that \mytitle~complements to existing token selection approaches. 
% \yin{Let us add a short sentence here to compare our best results with token selection w.r.t. base LLaVA-15, e.g., X folds reduction in FLOPs and y\% drop in accuracy.}
When integrated with Prumerge+ at an 85\% latency budget, our approach reduces computational requirements by 70\% while maintaining performance with only 1.7\% drop in accuracy.

%maintains competitive performance across multiple benchmarks while significantly reducing computational costs. 
% We observe that the performance impact varies between model variants: PruMerge experiences a more noticeable decline (-2.75\% at 85\% compute budget and -2.36\% at 60\%) compared to PruMerge+ (-2.23\% at 85\% compute budget and -1.66\% at 60\%). Both variants show larger performance drops under the same computational constraints compared to the original LLaVA-1.5 model. 
%For PruMerge+, we observe similar comparable results, achieves +2.45\% improvement at 85\% compute budget and only -1.01\% at 60\% compared to the original PruMerge+ full model performance.
%We observe \mytitle-H achieve a better performance on when integrating on Prumerge+ compared to integrating on the original LLaVA-1.5 model.
%For FastV, we observe similar results, with a more noticeable decline, where \mytitle-L achieves -2.35\% at 85\% compute budget and -3.7\% at 60\%, \mytitle-H achieves -0.05\% at 85\% compute budget and -4.1\% at 60\%.
%Surprisingly, we observe \mytitle-H achieves the near optimal performance at 85\% latency budget, when combining with token pruning techniques, sometimes outperform full model that uses token pruning.
%This suggests our approach-H performs particularly well when integrated with token pruning methods. Overall, our method serves as an effective complement to existing token selection approaches.

\subsection{Additional Analyses} 
\label{subsec:efficiency_results}

\begin{figure}[t]
    % First figure in a row
    % \begin{minipage}[t]{0.48\linewidth}
        \centering\vspace{-0.1em}
        {\includegraphics[width=0.85\linewidth]{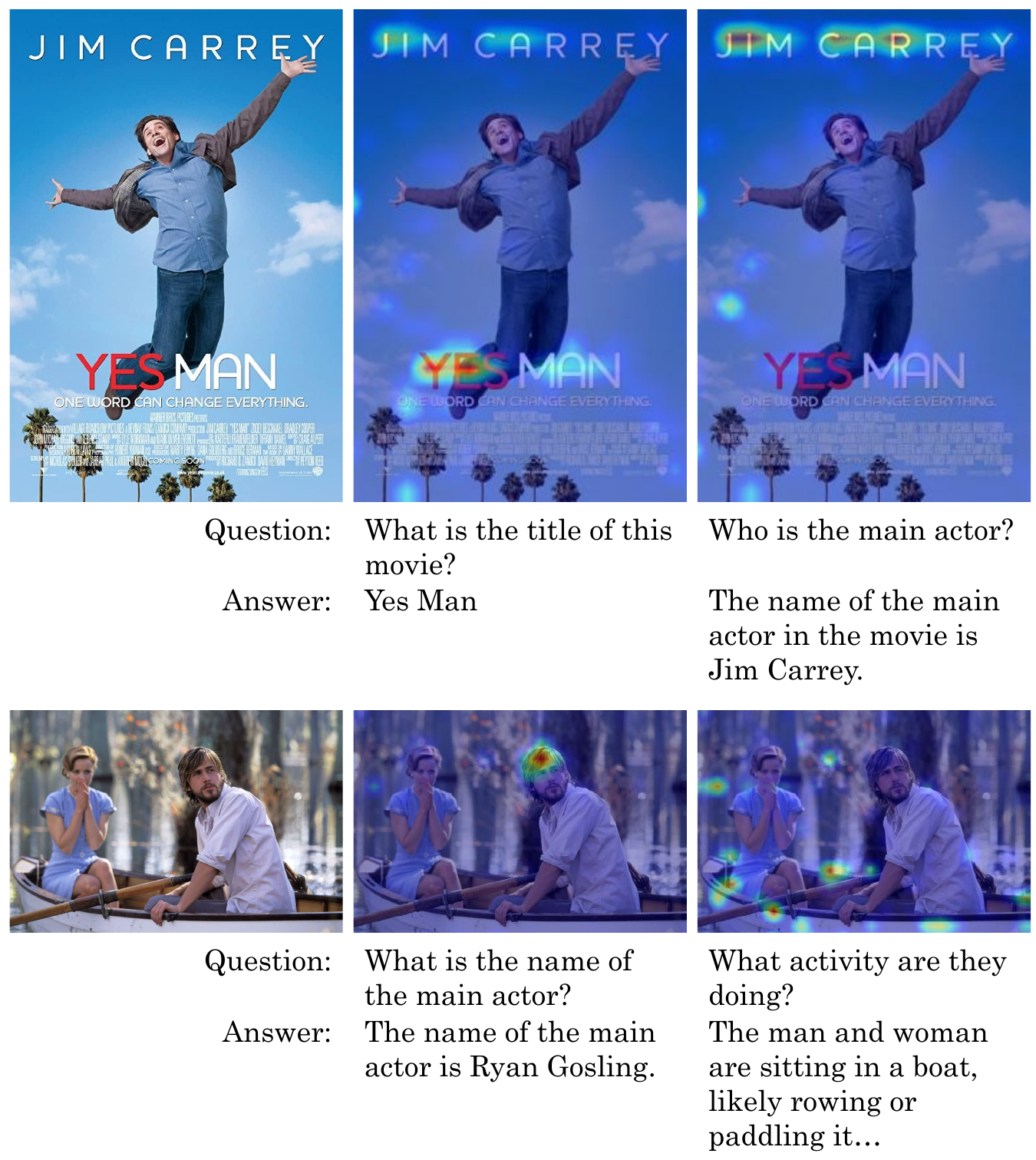}}\vspace{-0.7em}
        % \caption{Figure 1}
        % \label{fig:1}
    % \end{minipage}
    \caption{\textbf{Visualization of attention between the input latency token and visual tokens} with a 100\% latency budget.}
    \label{fig:attention_score_three_in_row}
    \vspace{-1.5em}
\end{figure}

\noindent \textbf{Latency adaptivity}.
We now evaluate the key capability of \mytitle: its adaptivity to input latency budget, \ie, \textit{the ability to complete inference under varying latency requirements using a single model}. We report the accuracy-latency tradeoff of \mytitle~variants (\ie, Pareto curves), both with and without token selection, on the VQAv2 benchmark. These results are shown in \cref{fig:binomial_7b_curve_with_fastv}.

Our results show that \mytitle~can empower a base MLLM with static compute footprint (\ie, LLaVA-1.5, PruMerge+, or FastV as individual dots in the \cref{fig:binomial_7b_curve_with_fastv}) to adapt to varying accuracy-latency tradeoffs (\ie, the corresponding curves in \cref{fig:binomial_7b_curve_with_fastv}). With varying latency budgets from 50\% to 100\%, \mytitle~effectively trades compute with accuracy. Integrating with token selection methods (PruMerge+ / FastV) further improves the overall efficiency. Thanks to our sampling process in the probabilistic modeling, \mytitle~maintains 0\% latency violation. We provide additional visualization of execution plans with different latency in \cref{fig:execution_plans_same_input} in \cref{supp:sec:latent_content_aware}. \smallskip

\begin{figure}[t]
    \centering
    % First figure in a row
    % \begin{minipage}[t]{0.48\linewidth}
        \centering\vspace{-0.1em}
        \fbox{\includegraphics[width=0.9\linewidth]{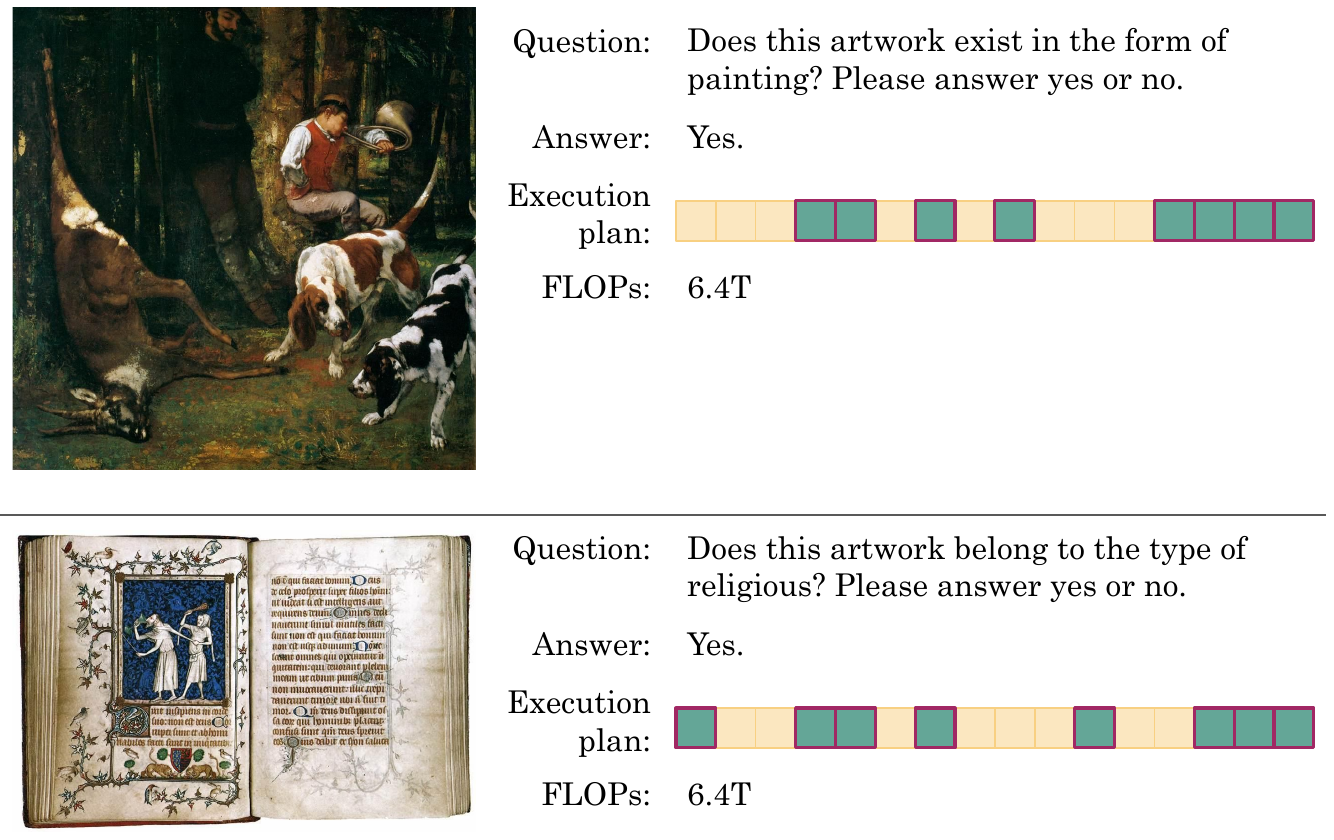}}\vspace{-0.5em}
        % \caption{Figure 1}
        % \label{fig:1}
    % \end{minipage}
    \caption{\textbf{Visualization of execution plans for different input}. The plan is color-coded with \textcolor[HTML]{64a697}{\textbf{enable}} or \textcolor[HTML]{f8d082}{\textbf{disable}} for the 16th to 32th Transformer blocks (left to right). The latency budget is 75\%.}
    \label{fig:execution_plans}
    \vspace{-1.5em}
\end{figure}
\noindent \textbf{Content adaptivity}.
It is worth noting that \mytitle~is also adaptive to the input content, \ie, \textit{with the same latency budget, its execution plan is dynamically adjusted based on input}. While not our main focus, we present results to illustrate our model's content adaptivity, with the aim of providing insights into its behavior and aiding in its diagnosis.

We visualize attention maps from the latency token to all input visual tokens, computed right before the latency token is fed into the scheduler. This is shown in \cref{fig:attention_score_three_in_row}. These attention maps highlight key regions in the input image for answering the target question. 
For example, in the top, attention concentrates on ``Yes Man'' for the movie title question but shifts to the actor name for actor identification question. %Similarly, in the bottom, attention focuses specifically on the character for actor identification question, yet spreads across the scene elements when describing activities. 
Further, we visualize the execution plans of different input content given by our scheduler in \cref{fig:execution_plans}. Under the same latency budget, \mytitle~generates distinct execution plans conditioned on the different visual content.
These results show \mytitle's ability to dynamically adjust its computational focus based on the input image and text query. See our \cref{supp:sec:latent_content_aware} for additional visualizations. \smallskip

\noindent \textbf{Generalization across MLLMs}.
We further demonstrate that \mytitle~can generalize to other MLLMs beyond LLaVA. We consider Mipha-3B~\cite{zhu2024comprehensive}, a lightweight MLLM built on Phi-2.7B~\cite{javaheripi2023phi}. Specifically, we apply \mytitle-L on Mipha-3B, following its training strategy~\cite{zhu2024comprehensive}, and report the results on MME benchmark, shown in \cref{tab:miphia}. These results have similar trend to those with LLaVA-1.5 in \cref{tab:main_table}. Complete results are presented in \cref{fig:mipha} in \cref{supp:sec:full_table}.\smallskip

%To demonstrate the generalizability of \mytitle, we conduct additional experiments on another MLLM named Mipha-3b \cite{zhu2024comprehensive}. Mipha-b is built on LLM Phi-2.7B \cite{javaheripi2023phi}. 
%We applied \mytitle-L  on Mipha-3b, following their training strategy. The results on MME, shown in \cref{fig:mipha}, are consistent with those using LLaVA-1.5 in the \cref{tab:main_table}. 

% Yin: the original table below. I removed the last two cols to make it consistent with the main tab.

\begin{comment}
\begin{table}[t!]
\centering
% \vspace{-0.14in}
\noindent \resizebox{\linewidth}{!}{
\begin{tabular}{lcccccccc}
    \toprule
    {Model} & VQAv2& SQA-IMG& TextVQA& POPE& MME& MMBench& GQA& MMVet\\
    \midrule
    Mipha-3B & 81.3& 70.9& 56.6& 86.7& 1488.9& 69.7& 63.9& 31.2\\
    w/ AdaLLaVA-L-100\%  & 81.1& 70.9& 55.3& 87.7& 1450.4& 69.2& 63.9& 34.2\\
    w/ AdaLLaVA-L-85\%  & 80.4& 71.0& 53.0& 87.8& 1429.3& 69.0& 63.9& 33.0 \\
    w/ AdaLLaVA-L-60\%  & 77.2& 68.4& 44.8& 88.0& 1397.3& 64.6& 61.2& 22.9 \\
    \bottomrule
\end{tabular}}
\caption{\textbf{Generalization} of \mytitle~to Mipha-3B.}
%We showcase the generalization of AdaLLaVA to another MLLM, namely Mipha-3B. Across 8 benchmarks, AdaLLaVA achieves nearly the same performance as the original model while reducing 15\% FLOPs.}
% \vspace{-0.2in}
\label{app:tab:miphia}
\end{table}
\end{comment}

\noindent \textbf{Ablation: probabilistic vs.\ deterministic modeling of the scheduler}. We present two design choices of the scheduler: deterministic and probabilistic (see \cref{subec:train_infer}). 
%The deterministic scheduler directly outputs execution plans and combines latency and language model losses. 
For our main results, we adopt the probabilistic version with conditional sampling (detailed in \cref{subsec:ModelInstantiation}). 
We now compare these two approaches across different latency budgets on the VQAv2 benchmark, using \mytitle-L 7B model. The results are summarized in \cref{table:modeling_comparison}. Our probabilistic model demonstrates superior adaptability across different latency budgets compared to the deterministic approach. We notice deterministic approach has noticeable performance drop given low latency budget due to under-utilization, and sometimes violates the latency budget. These results confirm our choice of the probabilistic modeling.\smallskip

%This suggests \mytitle-L achieves better resource efficiency while maintaining higher accuracy, particularly at stricter latency constraints.

\begin{table}[t!]
% \vspace{-0.14in}
\noindent \resizebox{\linewidth}{!}{
\centering
\begin{tabular}{lcccccc}
    \toprule
    {Model} & VQA\textsuperscript{v2}& SQA\textsuperscript{I}& VQA\textsuperscript{T}& POPE& MME& MMBench \\
    \midrule
    Mipha-3B & 81.3& 70.9& 56.6& 86.7& 1488.9& 69.7\\
    w/ AdaLLaVA-L-100\%  & 81.1& 70.9& 55.3& 87.7& 1450.4& 69.2\\
    w/ AdaLLaVA-L-85\%  & 80.4& 71.0& 53.0& 87.8& 1429.3& 69.0\\
    w/ AdaLLaVA-L-60\%  & 77.2& 68.4& 44.8& 88.0& 1397.3& 64.6\\
    \bottomrule
\end{tabular}}\vspace{-0.5em}
\caption{\textbf{Generalization} of \mytitle~to Mipha-3B.}
\label{tab:miphia}\vspace{-0.5em}
\end{table}

% \begin{figure}[h]
%     \centering
%     % First figure in a row
%     % \begin{minipage}[t]{0.48\linewidth}
%         \centering
%         {\includegraphics[width=\linewidth, trim = 0mm 0mm 0mm 0mm,clip]{figure/AdallavavsCombined.pdf}}
%         % \caption{Figure 1}
%         % \label{fig:1}
%     % \end{minipage}
%     \caption{The performance vs FLOPs usage given different latency budget. Combined-Loss: design latency loss and combine with original token loss by LLM. \mytitle: Conditional sampling based on the latency budget. \ZX{update legend, update the curve}}
%     \label{fig:combine_loss}
% \end{figure}

\begin{table}[t]
\centering
\resizebox{1.0\columnwidth}{!}{%
\begin{tabular}{c|c|c|c|c|c|c}
\toprule
 & \multicolumn{3}{c|}{\textbf{AdaLLaVA-L} (probabilistic scheduler)} & \multicolumn{3}{c}{\textbf{AdaLLaVA-L} (deterministic scheduler)} \\ 
 \midrule[0.5pt]
        {Latency budget}          & {Accuracy}       & {Success} (\%)       & Utilization (\%) & {Accuracy}       & {Success} (\%)      & Utilization (\%)\\ 
\midrule[0.5pt]
% 0.5          & 65.9           & 4.2            & 33.5           & 4.2           \\ 
% % \midrule[0.5pt]
% 0.56          & 68.6           & 4.8            & 62.9           & 4.5           \\ 
% % \midrule[0.5pt]
% 0.63          & 72.7           & 5.3            & 69.1           & 5.0           \\ 
% % \midrule[0.5pt]
% 0.69          & 73.6           & 5.8            & 70.2           & 5.3           \\ 
% % \midrule[0.5pt]
% 0.75          & 74.3           & 6.3            & 72.4           & 6.0           \\ 
% % \midrule[0.5pt]
% 0.81          & 75.0           & 6.9            & 75.4           & \textcolor{red}{8.2}           \\ 
% % \midrule[0.5pt]
% 0.88          & 75.2           & 7.4            & 75.5           & \textcolor{red}{8.5}           \\ 
% % \midrule[0.5pt]
% 0.94          & 75.6           & 7.9            & 75.7           & \textcolor{red}{8.5}           \\ 
% % \midrule[0.5pt]
% 1.0          & 76.2           & 8.5            & 75.0           & 8.5           \\ 

0.95 & 75.6 & 100.0 & 98.7 & 75.6 & 96.1  & 87.6 \\
0.85 & 74.9 & 100.0 & 99.2 & 74.6 & 100.0 & 80.4 \\
0.75 & 74.3 & 100.0 & 100.0 & 73.5 & 100.0 & 83.2 \\
0.65 & 72.7 & 100.0 & 96.5 & 72.2 & 100.0 & 83.1 \\
\bottomrule
\end{tabular}%
}\vspace{-0.5em}
\caption{\textbf{Ablation} on deterministic vs.\ probabilistic modeling for the scheduler. Results reported using 7B model on VQAv2.}\vspace{-1.2em}

%Results of \mytitle-L and deterministic scheduler across latency budget.}
% \textcolor{red}{Red} values indicate computation violation.}
\label{table:modeling_comparison}
% \vspace{-0.2in}
\end{table}

\noindent \textbf{Additional ablations}. Ablations on (1) the number and granularity of switches; (2) different designs of switches (\ie, \mytitle-H vs.\ \mytitle-L); and (3) sampling strategies are included in \cref{app:sec:ablation} due to space limit.

\section{Conclusion and Discussion}
In this paper, we introduced \mytitle, a novel adaptive inference framework designed for MLLMs. \mytitle~features a lightweight, learning-based scheduler and a probabilistic modeling technique. It empowers a base MLLM with the ability to adapt to varying latency budgets at inference time. Extensive experiments across benchmarks demonstrated that \mytitle~ is capable of producing latency- and content-aware execution plans, effectively achieving a range of accuracy-latency tradeoffs.\smallskip

%Moreover, our method is compatible with existing efficiency techniques, such as token pruning, further enhancing its practical utility. 

%We believe this work represents a step toward making MLLMs more viable for real-world applications where computational resources may fluctuate significantly.

\noindent \textbf{Adaptive inference of MLLMs}. Unlike LLMs, MLLMs include a vision encoder and process a large number of redundant visual tokens. While our paper focuses on the scheduling of the LLM component, this adaptivity can be further extended to token selection and vision encoder. 
% SB (8/1/25): Specifically we mean visual token selection. 
We hope this work will be a step toward making MLLMs more viable for real-world applications where computational resources may be constrained and fluctuate significantly.
%It should be noted that AdaLLaVA can be used for text-only LLMs, which presents an exciting direction and confirms the wide applicability of our approach.  
\smallskip

\noindent \textbf{Relationship to other efficiency methods}. This paper explores adaptive inference in MLLMs, emphasizing adaptability to varying latency budgets within a single model. Our approach is orthogonal to prior methods aimed at improving inference efficiency, such as sparse attention~\cite{child2019generating} and token selection~\cite{shang2024llava}. Indeed, many of these techniques (\eg token selection as shown in the paper) can be integrated with our framework to further enhance efficiency.\smallskip

\noindent \textbf{Practical deployment}. Our work focuses on algorithm-level innovation, leaving system-level optimization as future work. Conceptually, serving \mytitle~is similar to serving MoE-based LLMs~\cite{huang2024toward}, which also dynamically routes tokens to different execution paths based on the input. We express compute budgets as percentages of base model FLOPs to abstract hardware/software variations, leaving cross-device portability to future work.
% In our work, compute budgets are expressed as percentages of a base model’s FLOPs, as it abstracts away hardware / software specific variations. 
% We leave cross-device portability to future work. 
We invite joint effort from the vision, learning, and systems communities to further explore these directions.

% The following can go to the 9th page
\noindent \textbf{Acknowledgment}.
This research was supported in part by the National Science Foundation under Grant Numbers CNS 2333487 / 2333491 (CPS Frontier), CNS 2146449 (CAREER), and IIS 2442739 (CAREER), by the Army Research Lab under contract number W911NF-2020221, and by gift funding from Google and AWS. Any opinions, findings, and conclusions or recommendations expressed in this material are those of the authors and do not necessarily reflect the views of the sponsors.
%SC - looks good to me - thanks

{
    \bibliographystyle{ieeenat_fullname}
    \bibliography{ref}
}

\clearpage
\setcounter{figure}{0}
\setcounter{table}{0}
\setcounter{section}{0}
\setcounter{equation}{0}
\renewcommand{\thefigure}{\Alph{figure}}
\renewcommand{\thesection}{\Alph{section}}
\renewcommand{\thetable}{\Alph{table}}
\renewcommand{\theequation}{\Alph{equation}}
\setcounter{page}{1}

\maketitlesupplementary

\appendix

% \onecolumn
% \tableofcontents
% \twocolumn

In this supplementary material, we provide (1) additional implementation details (see \cref{supp:sec:Implementation_Details}); (2) detailed results, including broader benchmarks and other LLM backbones on \mytitle ~accompanying our experiments in \cref{sec:exps} (see \cref{supp:sec:full_table}); (3) further ablations on design of switches and comparison with naive sampling strategies (see \cref{app:sec:ablation}); (4) additional qualitative results on latency and content adaptivity (see \cref{supp:sec:latent_content_aware}) and (5) further discussion on practical deployment (see \cref{supp:sec:further_discussion}).
% and (5) additional attention map results on content adaptivity (see \cref{supp:sec:content_aware}). 
% and (4) provide further discussion of our work (see \cref{supp:sec:further_dis}). 
We hope that this document will complement our main paper. 

For sections, figures and equations, we use numbers (\eg, Sec.\ 1) to refer to the main paper and capital letters (\eg, Sec.\ A) to refer to this supplement.

\section{Further Implementation Details} \label{supp:sec:Implementation_Details}

\noindent \textbf{Probabilistic execution plan sampling}. Recall that in our probabilistic model, we define the distribution $p\left( \mathbf{s}|\{\mathbf{z}^{v|q}\}, l, \mathbf{\phi} \right)$ via a sampling process. Given the input tokens and a latency budget $l$, the output of the lightweight scheduler is a logits vector corresponding to the $K$ available switches: $\pi_1, \pi_2, \dots \pi_K \in \mathbb{R}$, where $\pi_i$ represents the relative likelihood of selecting the $i^{\text{th}}$ switch. The latency budget $l$ allows us to define $k$, the maximum number of switches allowed to activate. Then, a sampled execution plan from $p\left( \mathbf{s}|\{\mathbf{z}^{v|q}\}, l, \mathbf{\phi} \right)$ can be uniquely defined by a subset of $k$  distinct elements from $\{1, 2, \dots, K\}$, corresponding to its activated switches. We sample the execution plan by randomly picking $k$ switches one by one, without replacement, following the logits $\{\pi_i\}_{i=1}^K$. The complete sampling procedure is summarized in Algorithm~\ref{alg:sampling}, where Cat$\left(\Omega, \{\pi_i: i \in \Omega\}\right)$ denotes the categorical distribution of selecting an element from $\Omega$ with probabilities parameterized by $\{\eta_i: i \in \Omega\}$ = Softmax$(\{\pi_i: i \in \Omega\})$. The process ensures that the sampled execution plan adheres to the input budget while maximizing the utilization.

\begin{algorithm}
\caption{Sampling $\mathbf{s} \sim p\left( \mathbf{s}|\{\mathbf{z}^{v|q}\}, l, \mathbf{\phi} \right)$}\label{alg:sampling}
\begin{algorithmic}
\Require Latency budget $l$, sampling logits $\{\pi_i\}_{i=1}^K$
\Ensure Sampled binary vector $\mathbf{s}\in\{0,1\}^K$
\State Determine number of selections $k$ based on $l$
\State Initialize available set of switches $\Omega \gets \{1, 2, \dots, K\}$
\State Initialize $\mathbf{s} \gets (0, 0, \dots, 0) \in \{0,1\}^K$
\For{$i = 1$ to $k$}
\State Sample $\omega \sim$ Cat$\left(\Omega, \{\pi_i: i \in \Omega\}\right)$
\State $\mathbf{s}[\omega] \gets 1$ (activating the chosen switch)
\State $\Omega \gets \Omega \setminus \{\omega\}$
\EndFor
\Return $\mathbf{s}$
\end{algorithmic}
\end{algorithm}

\noindent \textbf{Differentiable sampling with Gumbel-Softmax}. Our designed scheduler is difficult to train as it involves a non-differentiable discrete sampling process, which prevents gradients from backpropagate to the scheduler during training. A common workaround involves using a score function estimator \cite{gu2015muprop, wu2018blockdrop}; however, this method often suffers from high variance and slow convergence. Instead, we employ Gumbel-Softmax~\cite{jang2017categorical}, a reparameterization trick for sampling from categorical distribution. In our implementation, the Gumbel-Softmax approximates $\omega \sim$ Cat$\left(\Omega, \{\pi_i: i \in \Omega\}\right)$ with a continuous random vector $\Tilde{\omega}$:
\begin{equation}
    \Tilde{\omega} = \text{Softmax} \left( [g_i + \log\eta_i] \right)_{i \in \Omega},
\end{equation}
where each $g_i$ is i.i.d. sample drawn from Gumbel$(0, 1)$; and $\eta_i = \text{Softmax}(\{\pi_j: j \in \Omega\})[i]$ is the probability of activating the $i^{\text{th}}$ switch, computed by the scheduler. Note that $\Tilde{\omega}$ is continuous and has a well-defined gradient. To maintain a hard execution plan, we take the one-hot encoding of $\Tilde{\omega}$ and apply the straight-through estimator (see~\cite{jang2017categorical} for more details).

\begin{figure}[t]
% \vspace{-2.2em}
\centering
\includegraphics[width=\linewidth]{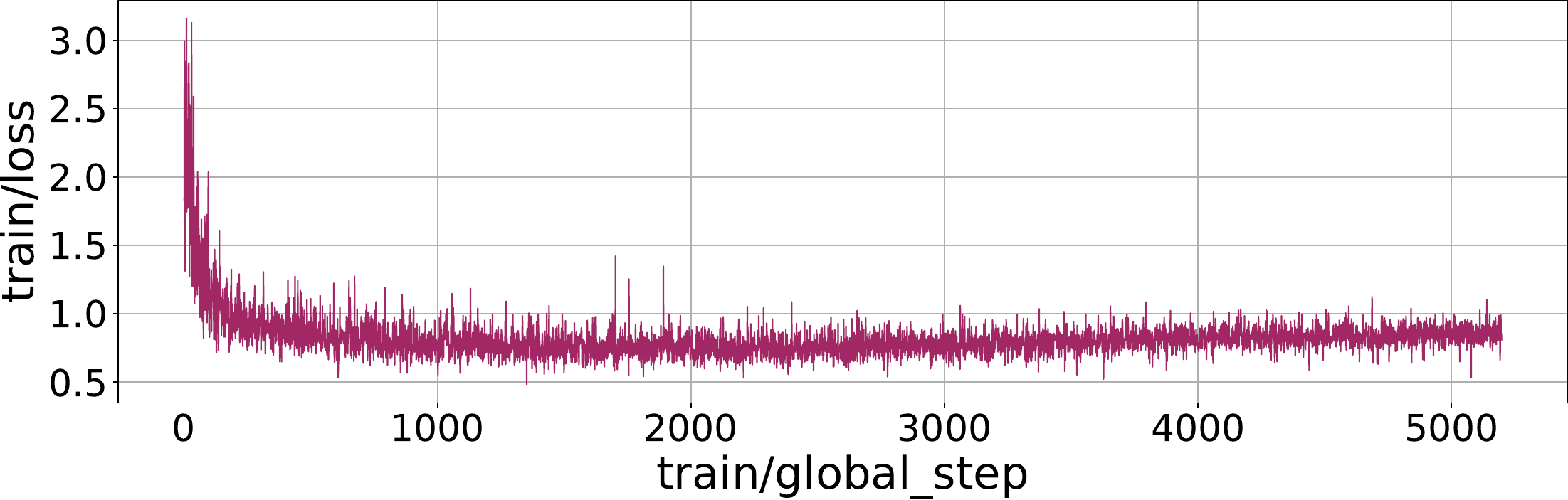}\vspace{-1em}
\caption{Training logs}
\label{fig:adallava-L_7b_train_log}
\vspace{-1.5em}
\end{figure}

\smallskip
\noindent \textbf{Training details}.
Training details were discussed in \cref{sec:exps} of the main paper. Here we show training curve of AdaLLaVA-L-7B with LLaVA 1.5 in \cref{fig:adallava-L_7b_train_log}.

\section{Detailed Results} \label{supp:sec:full_table}
\noindent \textbf{Full results on LLaVA 1.5}. We report the full set of results on LLaVA 1.5, LLaVA-PruMerge, LLaVA-PruMerge+ and FastV in \cref{tab:full_table}, as a complement to \cref{tab:main_table}. All experiments follow the same setting as described in \cref{subsec:exp_dataset_model}.  These results confirm that our \mytitle~framework successfully adapts to LLaVA 1.5 across different backbone sizes, and can be further combined with recent token selection methods (PruMerge, PruMerge+ and FastV) to further enhance efficiency. We maintain comparable performance while improving efficiency across multiple benchmarks. Additionally, our analysis reveals how performance varies under different latency constraints, demonstrating our framework's ability to trade between accuracy and latency. 

%We show full results 

\smallskip
\noindent \textbf{Broader benchmarks}. We extend our AdaLLaVA-L framework on broader benchmarks reported in \cite{liu2023improvedllava}, namely GQA~\cite{hudson2019gqa}, SEED-Bench~\cite{li2023seed}, MM-Vet~\cite{yu2023mm}, LLaVa-WILD~\cite{liu2023llava}, and VizWiz~\cite{gurari2018vizwiz} (see \cref{tab:more_task}). The model shows comparable performance and adaptive ability under different latency budget. The results demonstrate the strong generalization of \mytitle~to a wide range of benchmarks.

\begin{table}[t]
\centering
% \vspace{-0.12in}
\noindent \resizebox{\linewidth}{!}{
\begin{tabular}{lccccc}
    \toprule
    \bf{Model} & \bf{GQA} & \bf{SEED-Bench} & \bf{MM-Vet} & \bf{LLaVa-WILD} & \bf{VizWiz} \\
    \midrule
    LLaVA-1.5-7B        & 62.0 & 58.6 & 31.1 & 65.4 & 50.0 \\
    AdaLLaVA-L-7B-100\% & 61.5 & 60.5 & 30.7 & 64.2 & 54.3 \\
    AdaLLaVA-L-7B-85\%  & 61.3 & 60.2 & 30.0 & 62.1 & 51.5 \\
    AdaLLaVA-L-7B-60\%  & 58.7 & 59.8 & 23.9 & 46.3 & 44.8 \\
    \bottomrule
\end{tabular}}
% \vspace{-0.18in}
\caption{\textbf{Results on broader benchmarks}.}
\label{tab:more_task}
\end{table}

\mytitle-L maintains comparable performance under full computational budgets. With reduced compute budgets, \mytitle-L shows minimal performance degradation: an average accuracy drop of only 0.7\% at 85\% budget. Notably, \mytitle-L shows 1.5\% average performance boost at full compute budget.

\begin{figure}[t]
% \vspace{-0.2in}
\vspace{0.05in}
\begin{center}
{\includegraphics[width=0.8\linewidth]{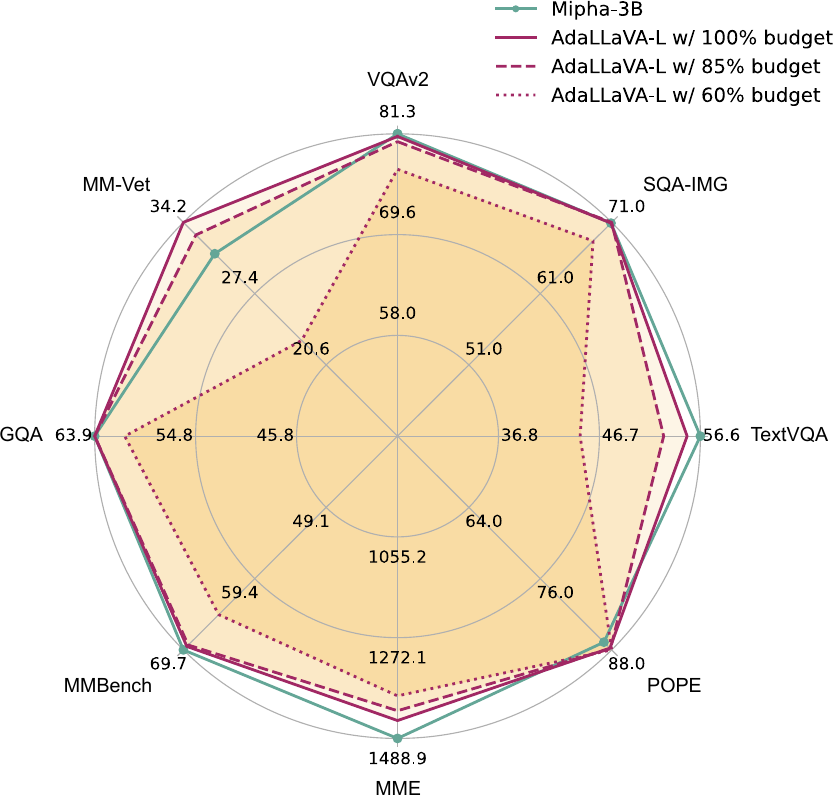}}
% \vspace{0.05in}
\end{center}
\vspace{-0.25in}
% \caption{We showcase the generalization of AdaLLaVA to another MLLM, namely Mipha-3B. Across 8 benchmarks, AdaLLaVA achieves nearly the same performance as the original model while reducing 15\% FLOPs.}
\caption{\textbf{Relative performance of applying AdaLLaVA-L to Mipha-3B under various latency budget}. The center of the radar corresponds to 60\% performance of the base Mipha-3B.}
\vspace{-1.5em}
\label{fig:mipha}
\end{figure}

\smallskip
\noindent \textbf{Generalization across MLLMs}.
We demonstrate that \mytitle~can generalize to other MLLMs beyond LLaVA. We consider Mipha-3B~\cite{zhu2024comprehensive}, a lightweight MLLM built on Phi-2.7B~\cite{javaheripi2023phi}. Specifically, we apply \mytitle-L on Mipha-3B, following its training strategy~\cite{zhu2024comprehensive}, and report the results on a comprehensive MLLM benchmark (MME), shown in \cref{fig:mipha}. We see that \mytitle-L maintains comparable performance under full computational budgets. With reduced compute budgets, \mytitle-L shows minimal performance degradation: an average accuracy drop of only 3.4\% at 85\% budget and 6.1\% at 60\% budget. These results have similar trend to those with LLaVA-1.5 in \cref{fig:teaser}.
% Yin: The following figure can be moved to supplement

%% full table
\begin{table*}[t]
\centering
\scalebox{0.86}{
\begin{tabular}{llccc|ccc|ccc}
\toprule
\multirow{2}{*}{Method} & \multirow{2}{*}{LLM} & Budget & FLOPs  & Prefill time  & VQA\textsuperscript{v2} & SQA\textsuperscript{I} & VQA\textsuperscript{T} & POPE & MME & MMBench \\
 & & (\%) & (T) & (ms) & \cite{goyal2017vqav2} & \cite{lu2022learn} & \cite{singh2019towards} & \cite{li2023pope} & \cite{fu2023mme} & \cite{liu2025mmbench}\\
\midrule
BLIP-2~\cite{li2023blip} & Vicuna-13B  & 100 & - & - & 41.0 & 61 & 42.5 & 85.3 & 1293.8 & - \\
InstructBLIP~\cite{dai2023instructblip} & Vicuna-7B & 100 & - & - & - & 60.5 & 50.1 & - & - & 36 \\
InstructBLIP~\cite{dai2023instructblip} & Vicuna-13B & 100 & - & - & - & 63.1 & 50.7 & 78.9 & 1212.8 & - \\
Shikra~\cite{chen2023shikra} & Vicuna-13B & 100 & - & - &  77.4 & - & - & - & - & 58.8 \\
IDEFICS-9B~\cite{IDEFICS} & LLaMA-7B & 100 & - & - &  50.9 & - & 25.9 & - & - & 48.2 \\
IDEFICS-80B~\cite{IDEFICS} & LLaMA-65B & 100 & - & - &  60.0 & - & 30.9 & - & - & 54.5 \\
Qwen-VL~\cite{Qwen-VL} & Qwen-7B & 100 & - & - & 78.8 & 67.1 & 63.8 & - & - & 38.2 \\
Qwen-VL-Chat~\cite{Qwen-VL} & Qwen-7B & 100 & - & - & 78.2 & 68.2 & 61.5 & - & 1487.5 & 60.6 \\
\hline
LLaVA-1.5~\cite{liu2023improvedllava} & Vicuna-7B & 100 & 8.6 & 81 & 78.5 & 66.8 & 58.2 & 85.9 & 1510.7 & 64.3 \\
\rowcolor{myPurple!30}
w/ \tmytitle \texttt{\small-L} & Vicuna-7B & 100 & 8.6 & 81 & 78.4 & 67.8 & 57.0 & 85.9 & 1521.0 &  63.7 \\
\rowcolor{myPurple!30}
w/ \tmytitle \texttt{\small-L} & Vicuna-7B & 85 & 7.2 & 69 & 77.1 & 67.4 & 54.5 & 86.4 & 1487.2 &  63.7 \\
\rowcolor{myPurple!30}
w/ \tmytitle \texttt{\small-L} & Vicuna-7B & 60 & 5.1 & 49 & 75.0 & 66.9 & 47.7 & 86.1 & 1463.8 & 63.8 \\
\rowcolor{myYellow!30}
w/ \tmytitle \texttt{\small-H} & Vicuna-7B & 100 & 8.6 & 81 & 77.9 & 68.5 & 57.1 & 86.9 & 1471.1 & 64.1 \\
\rowcolor{myYellow!30}
w/ \tmytitle \texttt{\small-H} & Vicuna-7B & 85 & 7.2 & 69 & 76.8 & 68.2 & 55.2 & 86.7 & 1494.9 & 64.3 \\
\rowcolor{myYellow!30}
w/ \tmytitle \texttt{\small-H} & Vicuna-7B & 60 & 5.1 & 49 & 74.2 & 68.1 & 48.7 & 85.0 & 1489.6 & 64.8 \\

\hline
LLaVA-1.5 & Vicuna-13B & 100 & 16.7 & 157 & 80.0 & 71.6 & 61.3 & 85.9 & 1531.3 & 67.7 \\
\rowcolor{myPurple!30}
w/ \tmytitle \texttt{\small-L} & Vicuna-13B & 100 & 16.7 & 157 & 79.7 & 72.4 & 59.9 & 86.9 & 1559.3 & 69.2 \\
\rowcolor{myPurple!30}
w/ \tmytitle \texttt{\small-L} & Vicuna-13B & 85 & 14.2 & 133 & 79.1 & 72.4 & 58.0 & 86.2 & 1563.9 & 68.9 \\
\rowcolor{myPurple!30}
w/ \tmytitle \texttt{\small-L} & Vicuna-13B & 60 & 10.0 & 94 & 77.4 & 71.8 & 54.3 & 87.3 & 1552.6 & 68.6 \\
%==================%==================%==================%================== prefix 16
\rowcolor{myYellow!30}
w/ \tmytitle \texttt{\small-H} & Vicuna-13B & 100 & 16.7 & 157 &  80.0 & 72.6 & 59.9 & 87.3 & 1531.9 & 67.4 \\
\rowcolor{myYellow!30}
w/ \tmytitle \texttt{\small-H} & Vicuna-13B & 85 & 14.2 & 133 & 78.9 & 72.3 & 59.0 & 86.1 & 1554.5 & 67.0 \\
\rowcolor{myYellow!30}
w/ \tmytitle \texttt{\small-H} & Vicuna-13B & 60 & 10.0 & 94 & 76.4 & 71.3 & 53.3 & 85.0 & 1529.5 & 66.9 \\
%==================%==================%==================%==================

\hline
Prumerge~\cite{shang2024llava} & Vicuna-7B & 100 & 1.4 & 16 & 72.0 & 68.5 & 56.0 & 76.3 & 1350.3 & 60.9 \\
\rowcolor{myPurple!30}
w/ \tmytitle \texttt{\small-L} & Vicuna-7B & 100 & 1.4 & 16 & 71.0 & 69.1 & 54.1 & 74.2 & 1312.6 & 58.4 \\
\rowcolor{myPurple!30}
w/ \tmytitle \texttt{\small-L} & Vicuna-7B & 85 & 1.2 & 14 & 69.7 & 68.6 & 52.5 & 75.6 & 1313.3 & 59.1 \\
\rowcolor{myPurple!30}
w/ \tmytitle \texttt{\small-L} & Vicuna-7B & 60 & 0.8 & 10 & 67.8 & 68.7 & 44.7 & 75.8 & 1332.5 & 57.0 \\
\rowcolor{myYellow!30}
w/ \tmytitle \texttt{\small-H} & Vicuna-7B & 100 & 1.4 & 16 & 70.4 & 67.9 & 54.4 & 77.2 & 1311.4 & 60.1 \\
\rowcolor{myYellow!30}
w/ \tmytitle \texttt{\small-H} & Vicuna-7B & 85 & 1.2 & 14 & 69.2 & 67.2 & 52.3 & 75.5 & 1309.7 & 60.7 \\
\rowcolor{myYellow!30}
w/ \tmytitle \texttt{\small-H}& Vicuna-7B & 60 & 0.8 & 10 & 66.8 & 68.1 & 45.9 & 76.4 & 1289.3 & 58.7 \\
\hline
Prumerge+ \cite{shang2024llava} & Vicuna-7B & 100 & 3.0 & 29 & 76.8 & 68.3 & 57.1 & 84.0 & 1462.4 & 64.9 \\ 
\rowcolor{myPurple!30}
w/ \tmytitle \texttt{\small-L} & Vicuna-7B & 100 & 3.0 & 29 & 76.3 & 68.3 & 55.8 & 85.1 & 1455.5 & 61.9 \\
\rowcolor{myPurple!30}
w/ \tmytitle \texttt{\small-L} & Vicuna-7B & 85 & 2.6 & 24 & 75.3 & 68.5 & 52.9 & 85.7 & 1429.5 & 62.5 \\
\rowcolor{myPurple!30}
w/ \tmytitle \texttt{\small-L} & Vicuna-7B & 60 & 1.8 & 17 & 73.0 & 67.7 & 47.4 & 85.6 & 1450.9 & 61.3 \\
\rowcolor{myYellow!30}
w/ \tmytitle \texttt{\small-H} & Vicuna-7B & 100 & 3.0 & 29 & 76.0 & 67.9 & 56.0 & 86.6 & 1503.2 & 63.2 \\
\rowcolor{myYellow!30}
w/ \tmytitle \texttt{\small-H}& Vicuna-7B & 85 & 2.6 & 24 & 75.0 & 68.1 & 54.2 & 86.4 & 1511.8 & 63.6 \\
\rowcolor{myYellow!30}
w/ \tmytitle \texttt{\small-H}& Vicuna-7B & 60 & 1.8 & 17 & 72.2 & 67.6 & 47.2 & 86.4 & 1458.0 & 63.6 \\
\hline
FastV (K=2,R=0.5)~\cite{chen2024image} & Vicuna-7B & 100 & 4.9 & 47 & 77.7 & 68.7 & 58.1 & 82.5 & 1516.2 & 64.3\\ 
\rowcolor{myPurple!30}
w/ \tmytitle \texttt{\small-L} & Vicuna-7B & 100 & 4.9 & 47 & 77.8 & 67.7 & 57.0 & 82.8 & 1494.3 & 63.5 \\
\rowcolor{myPurple!30}
w/ \tmytitle \texttt{\small-L} & Vicuna-7B & 85 & 4.2 & 40 & 76.9 & 67.8 & 54.4 & 83.3 & 1478.1 & 63.7\\
\rowcolor{myPurple!30}
w/ \tmytitle \texttt{\small-L} & Vicuna-7B & 60 & 3.0 & 29 & 74.5 & 67.0 & 47.2 & 83.8 & 1463.1 & 63.2\\
\rowcolor{myYellow!30}
w/ \tmytitle \texttt{\small-H} & Vicuna-7B & 100 & 4.9 & 47 & 77.4 & 68.4 & 57.0 & 84.3 & 1484.2 & 63.8 \\
\rowcolor{myYellow!30}
w/ \tmytitle \texttt{\small-H} & Vicuna-7B & 85 & 4.2 & 40 & 76.6 & 67.7 & 54.8 & 83.9 & 1520.5 & 63.9 \\
\rowcolor{myYellow!30}
w/ \tmytitle \texttt{\small-H} & Vicuna-7B & 60 & 3.0 & 29 & 73.9 & 68.3 & 48.7 & 82.4 & 1452.8 & 65.3 \\

% \hline
% Prumerge 13B & Vicuna-13B  & 100 & 1.8 & 30 & 72.8 & 71.0 & 58.4 & 78.5 & 1428.2 & 62.3 \\
% \rowcolor{myPurple!30}
% w/ \tmytitle \texttt{\small-L} & Vicuna-13B & 100 & 1.8 & 30 & 70.8 & 72.4 & 55.4 & 72.3 & 1343.5 & 60.7 \\
% \rowcolor{myPurple!30}
% w/ \tmytitle \texttt{\small-L} & Vicuna-13B & 85 & 1.5 & 26 & 70.4 & 72.7 & 55.3 & 71.5 & 1347.0 & 61.1 \\
% \rowcolor{myPurple!30}
% w/ \tmytitle \texttt{\small-L} & Vicuna-13B & 60 & 1.1 & 18 & 68.9 & 72.7 & 51.6 & 72.5 & 1350.7 & 60.0 \\
\bottomrule
\end{tabular}
}
\caption{\textbf{Results of MLLMs on six benchmarks}. Our \mytitle~ can be applied to LLaVA 1.5 with different size of LLM with different design of switches. Percentage (\%): The input latency requirement. \tmytitle\texttt{\small-L}: switches on selecting different transformer blocks. \tmytitle\texttt{\small-H}: switches on select different attention heads and MLP activations. VQA\textsuperscript{v2}: VQAv2 set. SQA\textsuperscript{I}: ScienceQA set. VQA\textsuperscript{T}: TextVQA set. Prumerge: LLaVA 1.5 with PruMerge.} 
\vspace{-0.2in}
\label{tab:full_table}
\end{table*}

\section{Additional Ablation Studies}  \label{app:sec:ablation}
% Yin: Move the following two ablations to the supplement.

% \subsection{Ablation Study} \label{subsec:ablation}
We now conduct ablation study, exploring different design choices. We explore the performance of different designs of tunable switches, namely \mytitle-L and \mytitle-H (detailed in \cref{subsec:ModelInstantiation}). 
% Due to space limit, we only present results with LLaVA 1.5-7B Model on VQAv2 dataset benchmark.
All results are reported with LLaVA 1.5-7B Model on VQAv2 dataset benchmark.

\begin{figure}[t]
    \centering
    % \vspace{-0.13in}
    \begin{minipage}{0.49\linewidth}
        \centering
        {\includegraphics[width=\linewidth]{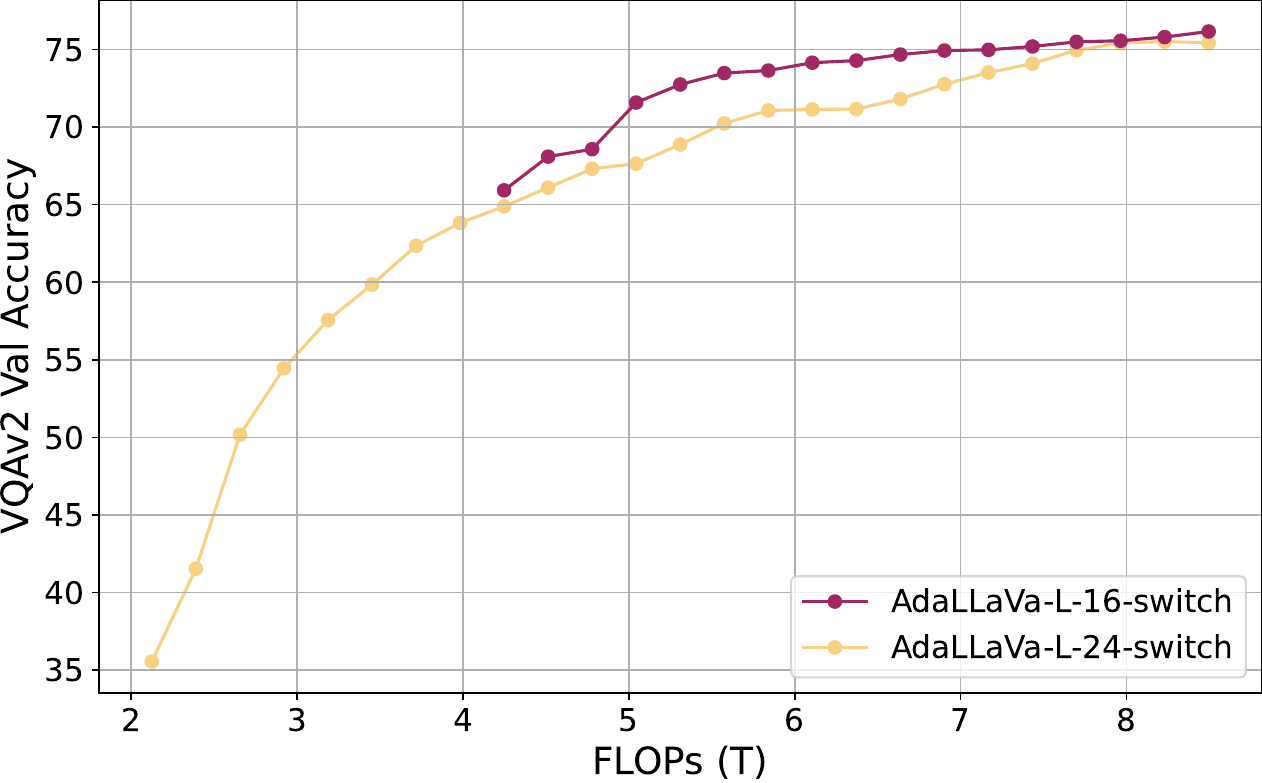}}
        \label{fig:L_curve_num_switch}
    \end{minipage}
    %\hfill
    \begin{minipage}{0.49\linewidth}
        \centering
        {\includegraphics[width=\linewidth]{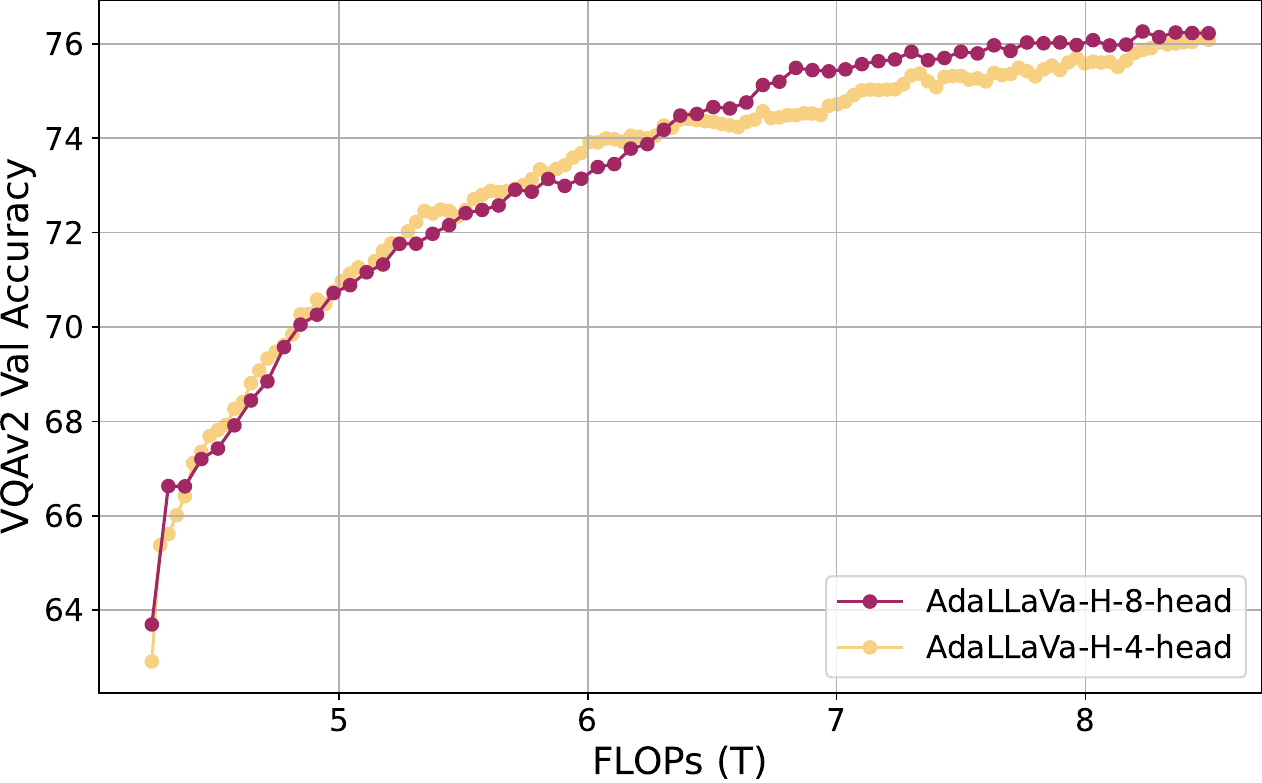}}
        \label{fig:H_curve_rank_switch}
    \end{minipage}
    \vspace{-1.5em}
    \caption{\textbf{Ablation studies on switch design choices}.}
    \vspace{-1.5em}
    \label{fig:curve_L_H}
\end{figure}

\smallskip
\noindent \textbf{Number \& granularity of switches}. 
We here conduct ablation studies to examine how the number and granularity of switches affect performance.
% \textit{Left}: increasing AdaLLaVA-L switches from 16 to 24 layers expands FLOPs control but severely degrades performance.
% Fig.~\ref{fig:ablation}(left) presents the results of increasing the number of switches in AdaLLaVA-L from 16 layers (as used in the main paper) to 24 layers.
% Although the 24-switch version provides a wider range of FLOPs control, it significantly degrades performance; whereas the 16-switch version achieves a better balance between efficiency and accuracy.
\cref{fig:curve_L_H} (\textbf{Left}) compares switches for the last 16 layers
(used in \cref{sec:exps}) versus 24 layers in \mytitle-L. While 24 switches enable finer FLOPs control, they significantly reduce model performance. The 16-switch configuration provides better accuracy while maintaining efficient adaptability.
% \textit{Right}: both 4- and 8-head AdaLLaVA-H groupings performs similarly, with 4 heads offering finer latency control.
% Fig.~\ref{fig:ablation}(right) compares AdaLLaVA-H with attention heads grouped in packs of 4 and 8 (used in the main paper). Both versions show similar performance and latency trade-offs, but the 4-head version offers finer control over latency.
\cref{fig:curve_L_H} (\textbf{Right}) evaluates attention sampling group sizes in \mytitle-H, focusing on operations within the last 16 layers. While both 4-head and 8-head (used in \cref{sec:exps}) configurations show comparable performance-latency tradeoffs, the 4-head version enables more granular latency control.

% \begin{figure}[t]
% % \vspace{-0.1in}
%     \centering
%     \begin{subfigure}{\linewidth}
%         \centering
%         \includegraphics[width=0.9\linewidth]{figure/rebuttal/binomial_7b.pdf}
%         \caption{Results on \mytitle-L 7b. 16/24: number of switches as 16/24.}
%         \label{fig:L_curve_num_switch}
%     \end{subfigure}
%     \vspace{2mm}  % Add some vertical space between subfigures
%     \begin{subfigure}{\linewidth}
%         \centering
%         \includegraphics[width=0.9\linewidth]{figure/rebuttal/binomial_7b-h.pdf}
%         \caption{Results on \mytitle-H 7b. 4/8: number of heads per group as 4/8.}
%         \label{fig:H_curve_rank_switch}
%     \end{subfigure}
%     \vspace{-0.3in}
%     \caption{\small Results on VQAv2 benchmark across latency budgets (FLOPs).}
%     \label{fig:curve_L_H}
%     \vspace{-0.2in}
% \end{figure}

\smallskip
\noindent \textbf{Design of the switches L vs H}.
We also explore the performance of design of tunable switches, particularly \mytitle-L versus \mytitle-H. Both methods allow adaptivity to latency requirements without significant modification to the pretrained LLM, while \mytitle-H offers better flexibility to latency input. 

As shown in \cref{fig:curve_L_H}, from FLOPs ranging from 5T to 8T, \mytitle-H-8-head shows slightly better performance overall, reaching approximately 76\% on VQA v2 Accuracy compared to \mytitle-L-16-switch which peaks around 75\%. 
% \cref{fig:HvsL} shows the performance scaling of our two switching strategies on VQAv2. 
% While \mytitle-L achieves slightly better accuracy across most computational budgets, 
Moreover, \mytitle-H demonstrates finer-grained control over the accuracy-latency trade-off. This is evident from the smoother curve of \mytitle-H, which can be attributed to its head/neuron-level switches providing more granular control over computational resources compared to the layer-level switches. This flexibility allows \mytitle-H to accommodate a wider range of latency budgets.
% though at a slight cost of lower peak performance.

% \input{figure/import_HvsL}

% \smallskip
% \noindent \textbf{Generalization across models}. %Yin: a short summary of our results cross board. 

\begin{figure}[t]
\centering
\includegraphics[width=0.95\linewidth]{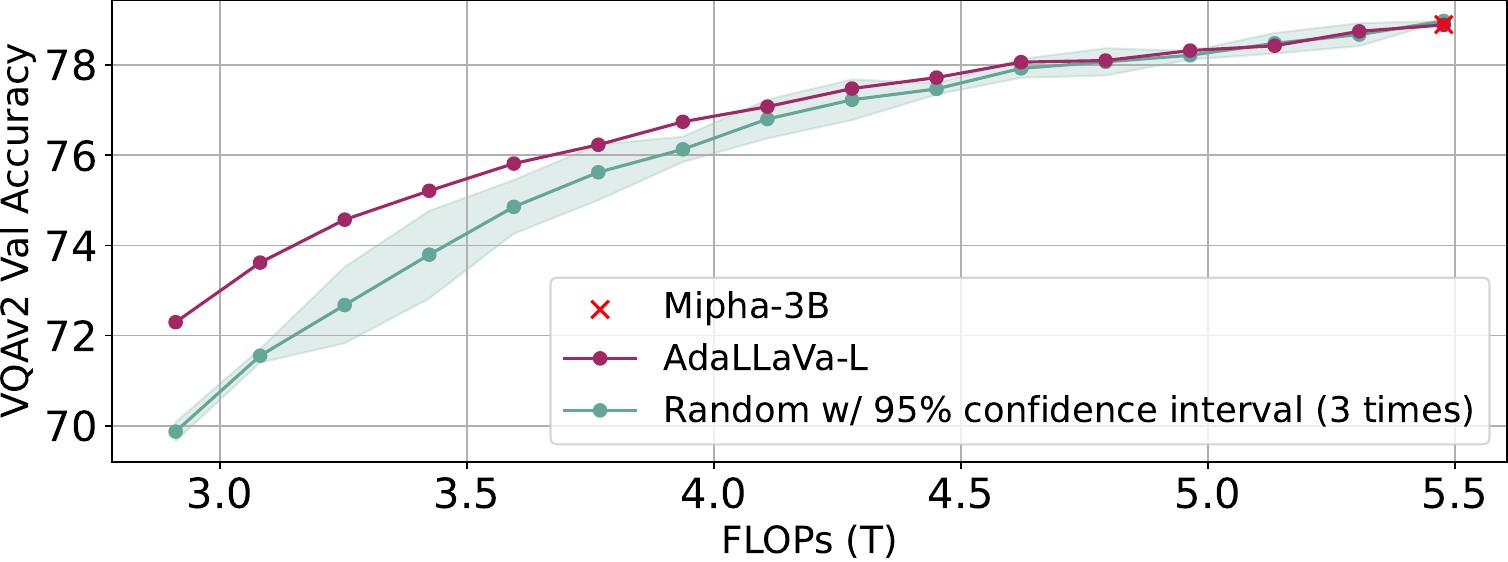}\vspace{-1em}
\caption{Comparison to random sampling}
\label{fig:random_sampling}
\vspace{-0.5em}
\end{figure}

\smallskip
\noindent \textbf{Comparison with naive sampling strategies.} 
We compare the performance of our AdaLLaVA-L versus random uniform sampling, where we disable the scheduler during training, showing in \cref{fig:random_sampling}. Both methods are built on Mipha-3B and fine-tuned using the same procedure. Random sampling is worse than AdaLLaVA and has high variance in results (shaded area).

\begin{figure}[t]
\centering
    \includegraphics[width=0.95\linewidth]{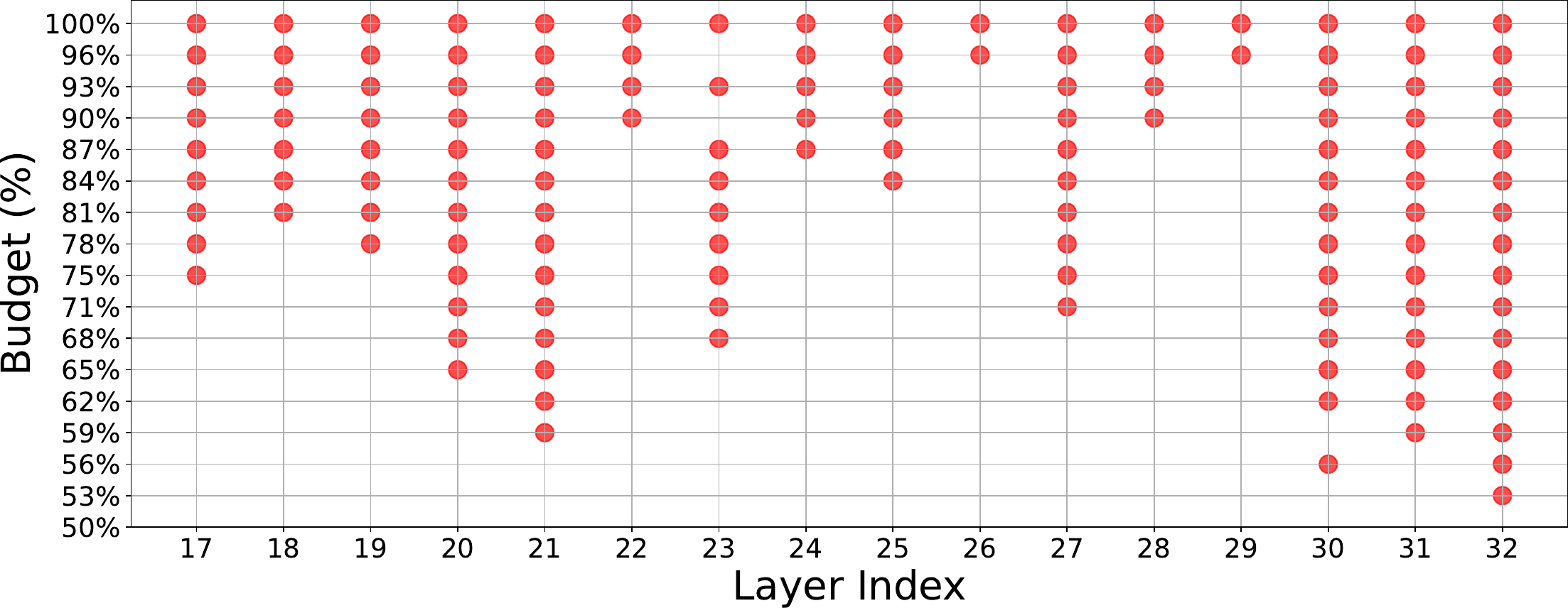}
    \vspace{-1em}
    \caption{Visualization of execution plan} 
\label{fig:execution_plans_same_input}
\vspace{-1.5em}
\end{figure}

\section{Additional Results on Adaptivity}
\label{supp:sec:latent_content_aware}
We provide further results to demonstrate \mytitle's latency and content adaptivity.

\smallskip
\noindent \textbf{Model Response under different latency}.
Here we show additional results on model response given same image-text input under different latency budget, similar to \cref{fig:teaser}. As shown in \cref{supp:tab:teaser}, given an image-query pair and latency constraint, \mytitle~ learns to generate appropriate responses while adapting to varying computational budgets.

\smallskip
\noindent \textbf{Visualization of execution plans with different latency}.
We report execution plans of the same input with varying budgets in \cref{fig:execution_plans_same_input}. As budget decreases, the scheduler prioritizes
keeping the last 3 layers over others. We have included outputs for the same input with varying budgets in \cref{supp:tab:teaser}.

\begin{table*}
  \begin{minipage}{0.99\linewidth}
\centering
\scalebox{0.78}{
\begin{tabular}{l p{5.5cm} p{5.5cm} p{5.5cm}}
\toprule
 \multicolumn{4}{l}{\bf Visual input example:}  \\
\midrule
&  \includegraphics[height=4.5cm]{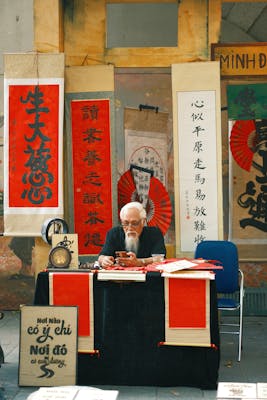} 
   & \includegraphics[height=3.5cm]{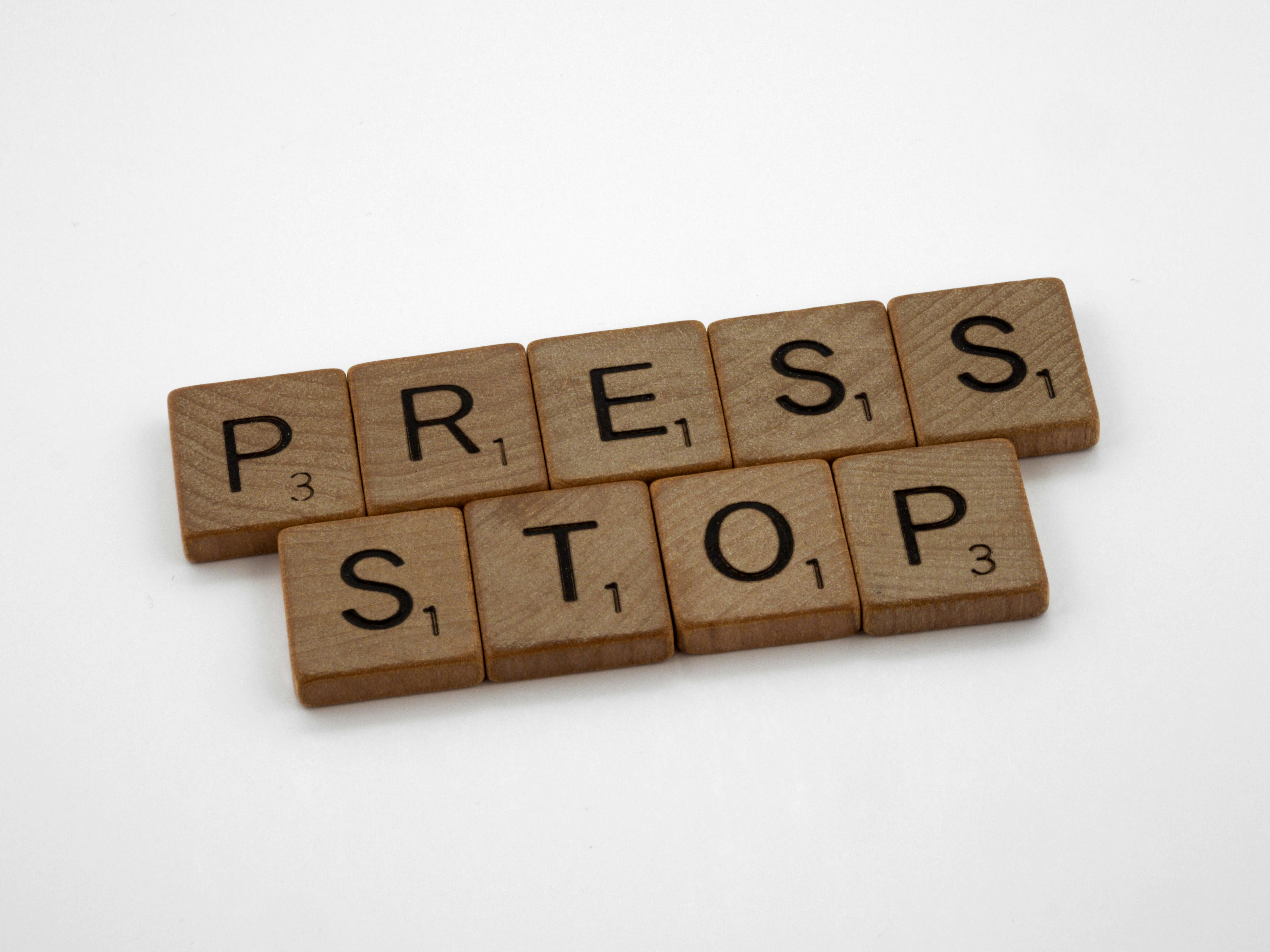}
   & \includegraphics[height=3.5cm]{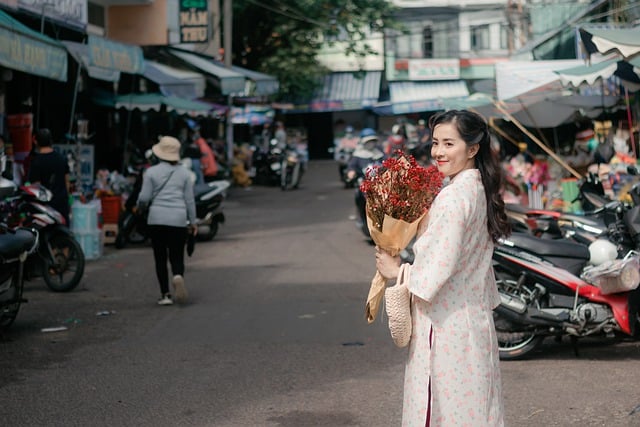} \\
User & Describe this image. 
     & Describe this image.
     & Describe this image. \\
\midrule
LLaVA 1.5-7B
&
The image features an older man sitting at a table, surrounded by various items. He is using a cell phone, possibly checking messages or browsing the internet...
& The image features a wooden block with a set of wooden letters arranged in a way that spells out the word "press stop." ...

& The image features a woman walking down a street, holding a bouquet of flowers in her hand...
\\ 
\midrule
\mytitle-L-50\% & The image depicts a man sitting at a table with a large number of papers and books on it. He is using a cell phone, possibly checking or communicating with someone. The man is wearing a tie, which adds to the professional setting.

The table is surrounded by several chairs, with one close to the man and others further away. A clock can be seen on the left side of the table, and a handbag is placed on the right side. The scene appears to be a workspace or a study area, with the man working on his tasks and using the cell phone.

& The image features a wooden board with a series of wooden chicks on it. The wooden chicks are placed in a row, with each one occupying a different position on the board. The arrangement of the chicks is a close-up of the word "press" and the word "stap.

& The image captures a young woman standing in a busy street, holding a bouquet of flowers. She appears to be smiling, possibly indicating that she is enjoying the moment. The street is filled with various vehicles, including several motorcycles and a motorcycle parked close to the woman.

There are numerous people in the scene, some standing closer to the woman, while others are further away. Some of the people are closer to the left side of the image, while others are further back, creating a sense of a busy city street. \\

\midrule
\mytitle-L-75\% & The image features a man sitting at a table, surrounded by various items. He is wearing a suit and tie, and appears to be using a cell phone. The table is set with a clock, a cup, and a book. There are also two chairs in the scene, one near the man and another further away.

In the background, there are two more books placed on the ground, and a handbag is located near the edge of the scene. The man seems to be engaged in a task or a conversation, as evidenced by his use of the cell phone.

& The image features a wooden word puzzle made up of four wooden letters, each with a number on them. The letters are arranged in a way that reads "Press Stop." The letters are placed on a white background, creating a visually appealing and playful display. The arrangement of the letters and numbers adds a unique and creative touch to the word puzzle.

& The image depicts a woman walking down a street, holding a bouquet of flowers in her hand. She is wearing a dress and appears to be enjoying her time outdoors. The street is lined with various vehicles, including motorcycles and a truck, parked along the side.

There are several people in the scene, some of whom are walking or standing near the vehicles. A handbag can be seen resting on the ground, possibly belonging to one of the individuals. The overall atmosphere of the scene is lively and bustling, with people going about their daily activities.
 \\
\midrule
\mytitle-L-100\% & The image features an older man sitting at a table, working on a laptop. He is surrounded by various items, including a cup, a cell phone, and a book. The table is adorned with a red and black tablecloth, and there are several signs and posters on the wall behind him.

In the room, there are two chairs, one near the man and another further away. A clock is also visible on the wall, indicating the time. The scene appears to be a workspace or a study area, with the man focused on his tasks.

& The image features a wooden tray with a set of wooden letters arranged on it. The letters spell out the word "Press Stop," with each letter positioned next to the other. The arrangement of the letters creates a visually appealing and organized display. The wooden tray itself is placed on a white surface, which further emphasizes the contrast between the wooden letters and the background.

& The image depicts a woman walking down a street, holding a bouquet of flowers in her hand. She is wearing a flowered dress and appears to be enjoying her time. The street is lined with various vehicles, including several motorcycles parked on both sides of the road.

There are also a few people walking around, some of whom are carrying handbags. The scene captures a lively atmosphere with people going about their daily activities.
 \\
\bottomrule
\end{tabular}
}
\captionof{table}{\textbf{\mytitle-L on LLaVA 1.5-7b model}, generating appropriate responses while adapting to varying computational budgets.}
\label{supp:tab:teaser}  
  \end{minipage}
\end{table*}

% \section{Additional Results on Content-Adaptivity}
% \label{supp:sec:content_aware}

% \subsection{Additional visualization for latency token attention}
\smallskip
\noindent \textbf{Visualization for latency token attention}.
We provide additional results on content awareness by showing the key-query attention scores of the
latency token and the input visual tokens with different text
questions, similar to \cref{fig:attention_score_three_in_row}.

\begin{figure}[t]
    \centering
    % First figure in a row
    % \begin{minipage}[t]{0.48\linewidth}
        \centering
        {\includegraphics[width=0.9\linewidth, trim = 2mm 0mm 30mm 0mm,clip]{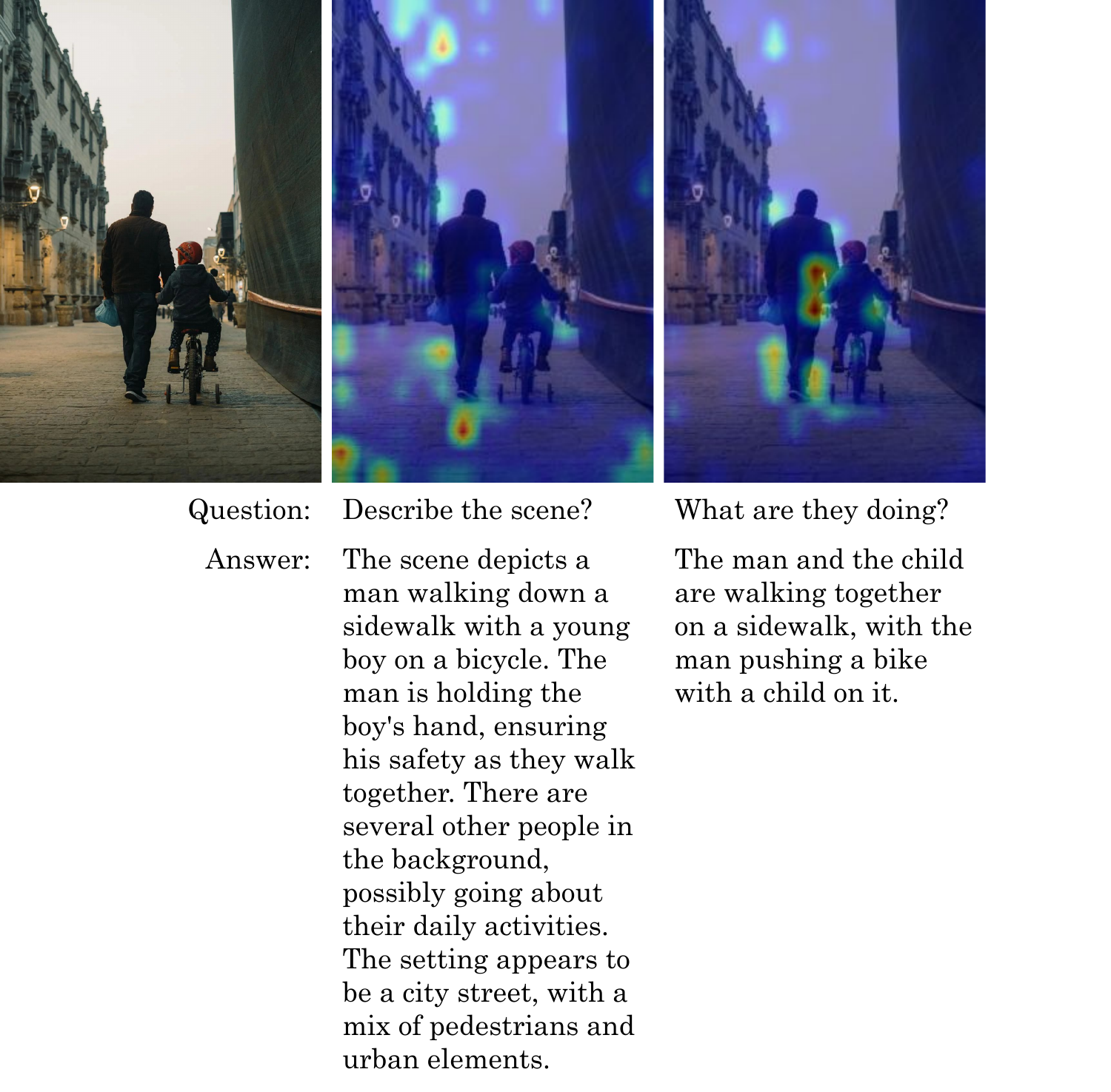}}
        {\includegraphics[width=0.9\linewidth, trim = 2mm 111mm 30mm 0mm,clip]{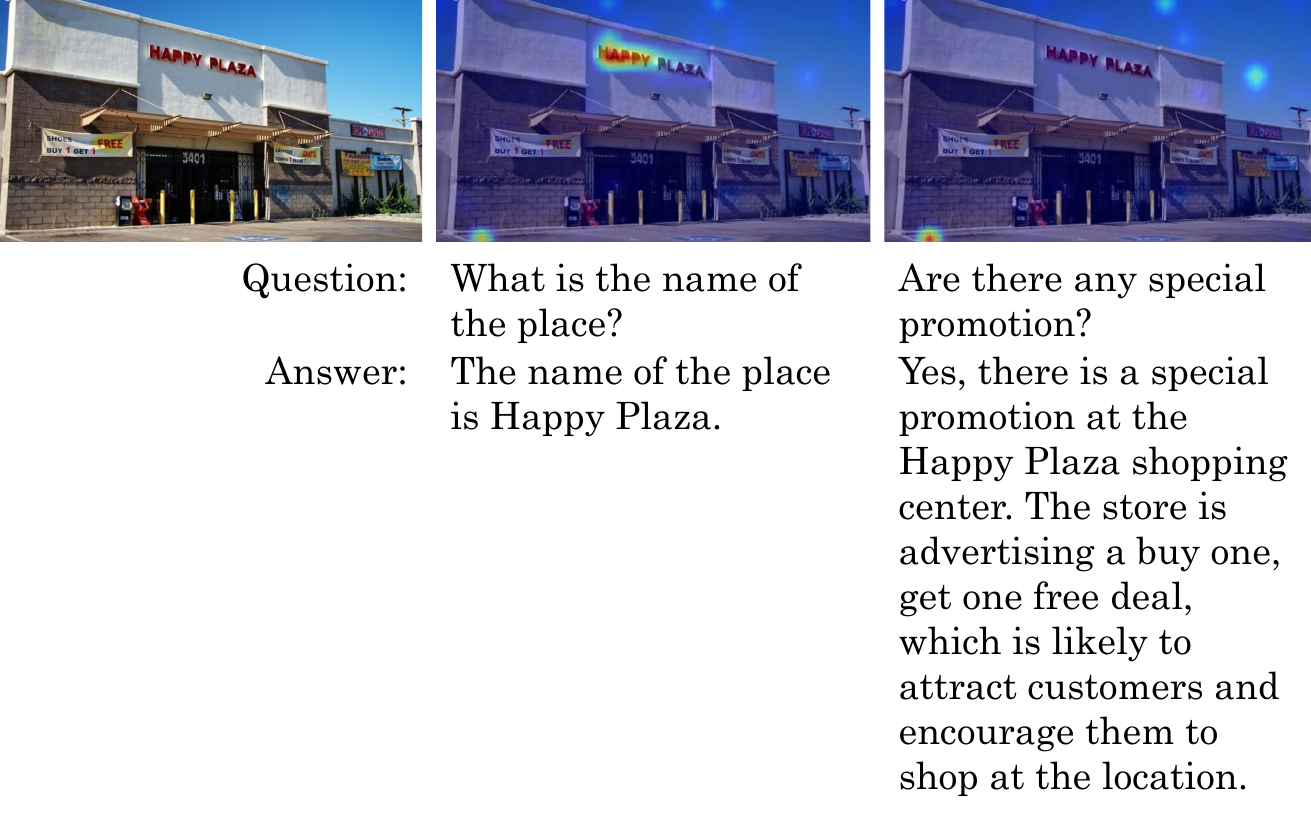}}
        {\includegraphics[width=0.9\linewidth, trim = 2mm 43mm 30mm 0mm,clip]{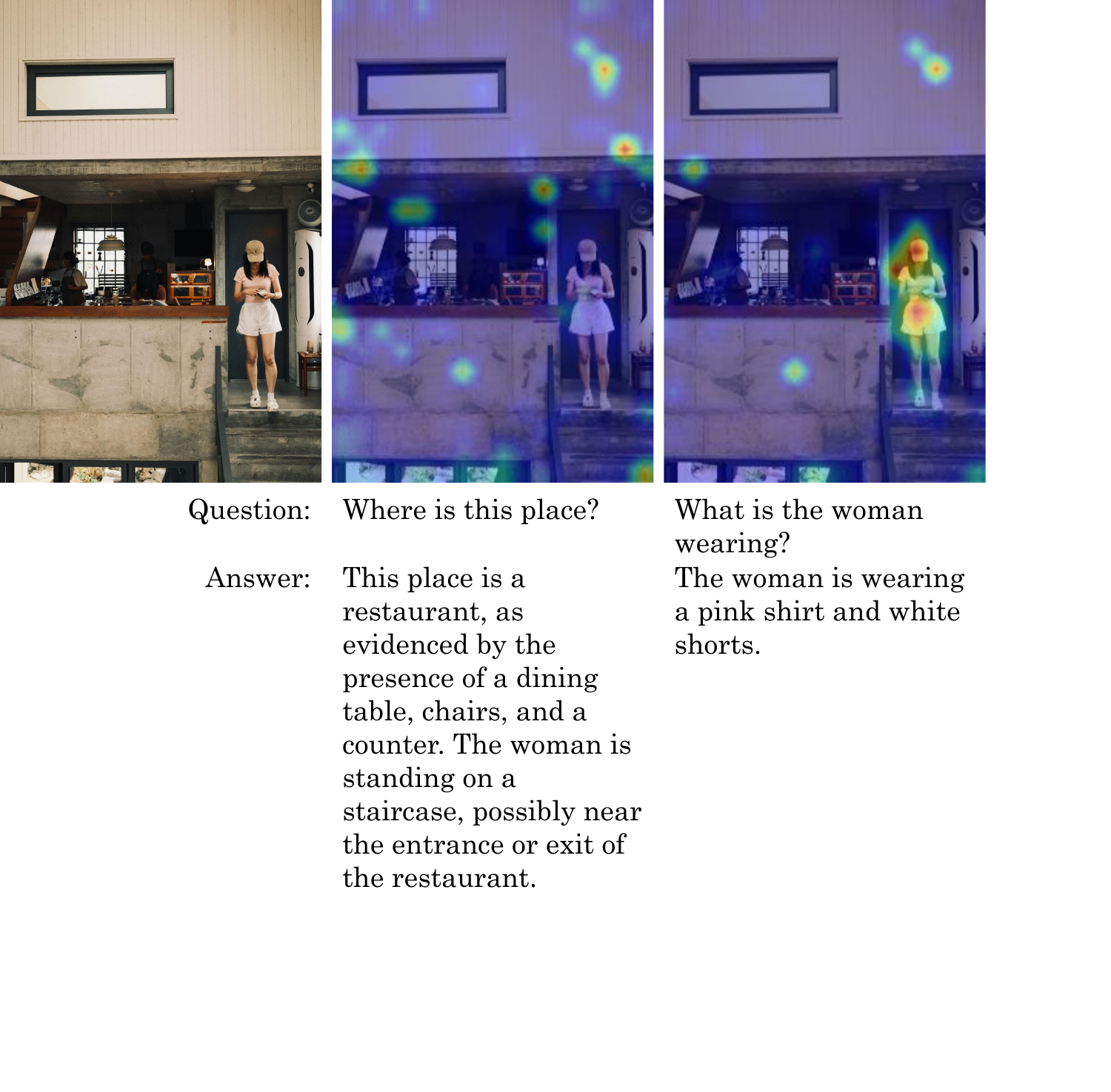}}
        % \caption{Figure 1}
        % \label{fig:1}
    % \end{minipage}
    \vspace{-0.8em}
    \caption{The key-query attention scores between latency token and visual tokens. The latency input is 1.0 in these examples.}
    \label{supp:fig:attention_score_three_in_row}
    \vspace{-0.2in}
\end{figure}

\cref{supp:fig:attention_score_three_in_row} further demonstrate the model's content-aware behavior. 
In the father-child scene image, attention spans the entire street for scene description but focuses centrally for query asking for activity. For Happy Plaza image, attention targets the storefront sign for location queries but shifts to promotional areas for query about special offers. In the restaurant scene, attention distributes across interior elements when identifying location type but concentrates on the woman's clothing for attire questions. This consistently shows model adjusts its attention based on the query.

% For in the father-child scene, attention spreads across the entire street view for scene description but concentrates on the middle when asking about their activity.
% For the Happy Plaza image, attention focuses on the storefront sign when asking about the location name, but shifts to the promotional signage area when querying about special offers. Similarly, in the restaurant scene, attention distributes across the interior elements (tables, counter, chairs) when identifying the location type, but concentrates specifically on the woman's clothing when asked about her attire. 
% This consistently shows how the model adjusts its attention based on query requirements.

\smallskip
% \subsection{Evolution of latency token across layers.} 
\noindent \textbf{Visualization for latency token across layers}.
We plot the evolution of the latency token from layers 12 to 16 using the same example in \cref{fig:attention_score_three_in_row} of the main paper (see figure below).
As seen in \cref{supp:fig:evolution_latency}, the latency token progressively gathers key information from the input visual tokens for scheduling. 

\begin{figure}[H]
\begin{center}
% \vspace{-0.8em}
\includegraphics[width=0.95\linewidth, trim = 0mm 176mm 0mm 0mm]{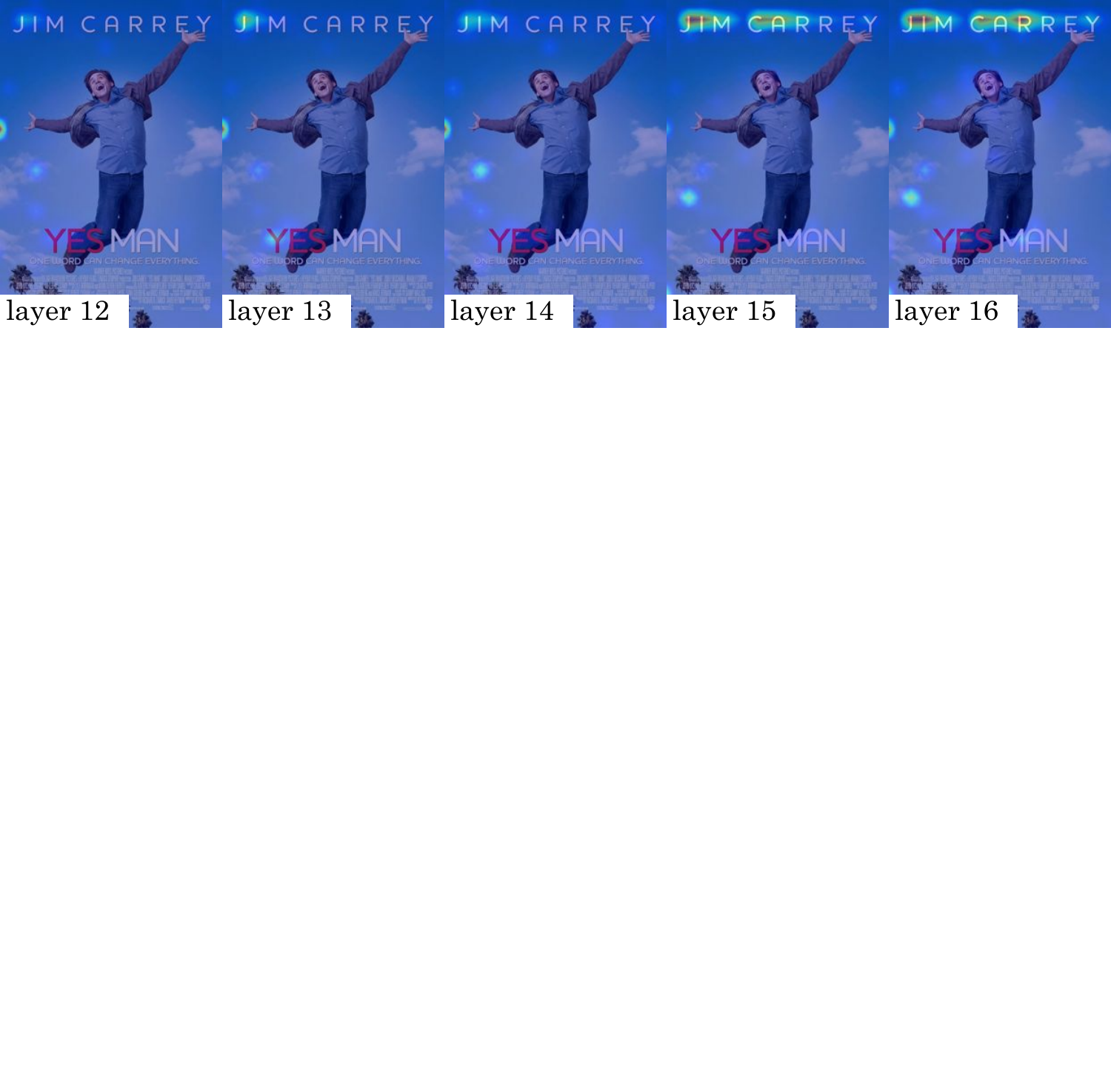}

\caption{Evolution of latency token across layers in AdaLLaVA-L on 7b model.}
\label{supp:fig:evolution_latency}
\end{center}
\vspace{-1em}
\end{figure}

\section{Further Discussion} \label{supp:sec:further_discussion}
\noindent \textbf{FLOPs, latency, and cross-device portability.}
In our work, compute budgets are expressed as percentages of a base model’s FLOPs. Percentages can be translated into absolute FLOP targets, given base model’s architecture and input size, from which total FLOPs can be estimated. We choose FLOPs over runtime latency, as it abstracts away hardware / software specific variations
%, allowing us to focus on algorithmic design
. We acknowledge that the relationship between FLOPs and latency vary across devices, depending on model architecture, hardware and software. We leave cross-device portability to future work.

\clearpage

\end{document}